\documentclass[11pt,twoside]{article}
\usepackage[preprint]{jmlr2e}

\usepackage{graphicx}   % for including figures
\usepackage{amsmath, amssymb}
\usepackage{booktabs}   % nice tables
\usepackage{hyperref}   % hyperlinks
\usepackage{caption}
\usepackage{subcaption} % for subfigures
\usepackage{xcolor}
\usepackage{enumitem}
\usepackage{url}
\usepackage[linesnumbered,ruled,vlined]{algorithm2e}
\usepackage{verbatimbox}
\usepackage{framed}
\usepackage{tabularx}

\graphicspath{{figures/}}

\title{UpBench: A Dynamically Evolving Real-World Labor-Market Agentic Benchmark Framework Built for Human-Centric AI}

\author{
    \name \hspace*{-0.4em}Darvin Yi \email darvinyi@upwork.com \\
    \AND
    \name Teng Liu \email tedliu@upwork.com \\
    \AND
    \name Mattie Terzolo \email mattieterzolo@upwork.com\\
    \AND
    \name Lance Hasson \email lancehasson@upwork.com\\
    \AND
    \name Ayan Sinha \email ayansinha@upwork.com\\
    \AND
    \name Pablo Mendes \email pablomendes@upwork.com\\
    \AND
    \name Andrew Rabinovich \email andrewrabinovich@upwork.com
}

\ShortHeadings{UpBench}{Yi et al.}
\firstpageno{1}

\begin{document}

\maketitle

\begin{abstract}
As large language model (LLM) agents increasingly undertake digital work, reliable frameworks are needed to evaluate their real-world competence, adaptability, and capacity for human collaboration. Existing benchmarks remain largely static, synthetic, or domain-limited, providing limited insight into how agents perform in dynamic, economically meaningful environments. We introduce UpBench, a dynamically evolving benchmark grounded in real jobs drawn from the global Upwork labor marketplace. Each task corresponds to a verified client transaction, anchoring evaluation in genuine work activity and financial outcomes. UpBench employs a rubric-based evaluation framework, in which expert freelancers decompose each job into detailed, verifiable acceptance criteria and assess AI submissions with per-criterion feedback. This structure enables fine-grained analysis of model strengths, weaknesses, and instruction-following fidelity beyond binary pass/fail metrics. Human expertise is integrated throughout the data pipeline (from job curation and rubric construction to evaluation) ensuring fidelity to real professional standards and supporting research on human-AI collaboration. By regularly refreshing tasks to reflect the evolving nature of online work, UpBench provides a scalable, human-centered foundation for evaluating agentic systems in authentic labor-market contexts, offering a path toward a collaborative framework,  where AI amplifies human capability through partnership rather than replacement.
\end{abstract}


% --------------------------------------------------
% Sections
% --------------------------------------------------
\section{Introduction}

Large language model (LLM) agents are increasingly capable of reasoning, planning, and acting in complex environments. As these systems begin to perform more sustained digital work,
evaluation methods must evolve accordingly. Traditional benchmarks are static, synthetic, and narrowly scoped, which offer limited insight into how agents perform on tasks that reflect real professional expectations. Progress in agentic systems therefore requires benchmarks that capture the dynamics, variability, and feedback-rich structure of real-world work.

We introduce \textbf{UpBench}, a large-scale and continually evolving benchmark designed to evaluate agent performance on economically grounded tasks. Each example in UpBench corresponds to a real job drawn from a global labor marketplace, including historical deliverables and verified client payouts. This foundation anchors evaluation in genuine human demand and practical utility rather than synthetic proxies.

UpBench emphasizes both breadth and depth. The dataset spans diverse professional domains and includes detailed rubrics that decompose tasks into verifiable acceptance criteria. Each AI submission is paired with per-criterion feedback generated by human experts, enabling fine-grained analysis of model behavior beyond binary pass/fail metrics. Because the dataset is refreshed over time, it also reflects the evolving structure of work and skills in the digital economy.

Beyond its use as a benchmark, UpBench provides an environment for studying human–AI collaboration. The data pipeline integrates human expertise at multiple stages (i.e., qualification, rubric creation, and evaluation) allowing systematic examination of how human feedback shapes agent performance and alignment. This human-in-the-loop design supports research on co-evolutionary training, supervision quality, and the role of human judgment in agent development.

Together, these features position UpBench as a scalable foundation for evaluating and improving agentic systems in realistic contexts. By grounding assessment in real economic activity and human feedback, it provides a path toward measuring how AI agents can augment, rather than simply automate, human work.
\section{Related Work}

\subsection{Agentic AI Benchmarks} 

A growing body of work recognizes that evaluating AI agents requires moving beyond traditional static tasks to agentic settings in which systems plan, act, and deliver outcomes in more realistic work-like environments. For example, the recent benchmark described in “Holistic Agent Leaderboard: The Missing Infrastructure for AI Agent Evaluation” addresses “holistic agent” evaluation across nine models and nine benchmark domains (coding, web navigation, science, customer service), reporting over 21,000 rollouts to analyze interactions among model, scaffold, and benchmark (\cite{kapoor2025holistic}).

Other general-purpose agent benchmarks such as “AgentBench: Evaluating LLMs as Agents” explore LLM-agent performance in interactive, multi-step decision-making environments (\cite{liu2023agentbench}). More recently, frameworks like “Establishing Best Practices for Building Rigorous Agentic Benchmarks” have identified methodological gaps in benchmark design (hold-out sets, reward design, task specification) and offered a checklist for more rigorous agentic evaluation (\cite{zhu2025establishing}).

In the sub-domain of economically grounded work evaluation, two notable initiatives are worth highlighting. First, the AI Productivity Index (APEX) from Mercor assesses frontier AI models on tasks drawn from investment banking, consulting, law, and medicine (i.e., knowledge-work sectors with direct economic relevance) (\cite{vidgen2025ai}). Second, GDPval from OpenAI evaluates model performance on tasks from 44 occupations covering major GDP-contributing sectors, with deliverables mapped to real-world work (\cite{patwardhan2025gdpval}).

These efforts signal a shift toward benchmarking what agents can do in person-work settings, rather than purely synthetic reasoning or code-completion tasks. However, several limitations remain: many benchmarks focus on single-turn deliverables rather than full workflows, may use proxies rather than true client-facing deliverables, and often lack ongoing temporal refresh or rubrics with per-criterion feedback.

\subsection{Human-in-the-Loop \& Centaur Evaluation Studies}

Beyond benchmark design, there is a rich literature on how humans are integrated into evaluation loops—both as collaborators with agents (centaur settings) and as supervisors of agent behaviour. The position paper AI Should Not Be an Imitation Game: Centaur Evaluations argues that evaluation regimes should shift from pure automation (agent alone) to centaur evaluations, in which humans and AI systems jointly solve tasks, enabling measurement of human augmentation, interpretability, and collaboration (\cite{hauptposition}). Studies of human-in-the-loop systems in machine learning more broadly show, for example, how allowing human feedback may reduce trust in a system even if accuracy improves (\cite{honeycutt2020soliciting}). These works underscore the importance of designing pipelines that respect human-agent interaction, supervision quality, and feedback granularity—not just agent autonomy. For agentic work benchmarks, this means tasks, rubrics, feedback loops, and human–AI orchestration become evaluation design elements.

\subsection{Previous Benchmarks Built on Upwork Data}

This paper does not represent the first time Upwork's Labormarket data was used in the construction of an agentic benchmark.  SWE-Lancer (\cite{miserendino2025swelancerfrontierllmsearn}) evaluates frontier LLMs on real freelance software-engineering projects, measuring their ability to complete tasks that previously resulted in actual client payments. It provides one of the first economically grounded benchmarks for end-to-end coding work.  Additionally, the Remote Labor Index (RLI) is a multi‑sector benchmark built from real, paid remote‑work projects, sourced largely from freelance platforms and organized using the Upwork taxonomy.  The RLI aimed to evaluate end‑to‑end AI agents on economically valuable tasks with full briefs, inputs, and human gold‑standard deliverables (\cite{mazeika2025remote}).

Additionally, our prior work lays the foundation for the current study. In a prior paper, Towards Real‑World Evaluation of Agentic Work in Freelance Marketplaces we introduced a dataset derived from real freelance marketplace jobs, grounding tasks in actual economic transactions. Secondly, in A Scalable Agentic Environment for Real World Work we expanded the pipeline to support multi-agent interaction and a structured data-collection architecture. These contributions established (i) the viability of real job-based benchmarks, (ii) an initial rubric and feedback system, and (iii) a human–AI collaboration pipeline (\cite{terzolo2025upbench}).
The present work builds on those foundations by scaling dataset size, increasing domain diversity, deepening rubric granularity with per-criterion feedback, and introducing a continuously refreshed stream of tasks.

\subsection{Our Contributions}

Relative to the existing body of work, our paper makes the following major contributions: 1) We use real-world jobs where every task corresponds to a historical job with verified client payout, affording economic grounding and authenticity. 2) We create hand-generated rubrics attached to each job, decomposing requirements into acceptance criteria. 3) We implement human grading of rubrics with feedback, where expert human freelancers grade model submissions along each rubric criterion and provide per-criterion feedback, enabling fine-grained evaluation. 4) We establish a human-centric data-collection and evaluation pipeline that, rather than only evaluating agents, emphasizes human–AI collaboration in data collection, qualification, rubric design, feedback loops and continuous refresh of tasks.
\section{Method}

Our data pipeline transforms raw marketplace records into a research-grade benchmark suitable for large-scale evaluation of AI agents on real professional work. The pipeline consists of four main components: filtering jobs, additional job categorization, rubric creation, and evaluation of job deliverables.

\subsection{Human-Centric Data Collection}

A central feature of the UpBench pipeline is its reliance on human expertise for dataset construction, annotation, and evaluation. Rather than relying solely on synthetic supervision or crowd-sourced labeling, we engage verified experts from Upwork’s own freelancer network to ensure that each stage of the data-generation process reflects real professional standards.

All contributing experts are active Upwork freelancers who meet strict eligibility criteria. Every recruited expert freelancer has a job satisfaction score of 100\% with every freelancer having a top rated or top rated plus badge.  Our freelancers have cumulative earnings of over \$1M on the Upwork platform with a combined 96,000 hours on the platform. Experts are contracted directly through Upwork, allowing the data-collection process to leverage the same professional marketplace dynamics that underlie the benchmark itself.

Each expert participates in one or more stages of data collection, including qualification verification, rubric construction, and deliverable evaluation. Jobs are distributed among experts according to their domain specialization, corresponding to eight predefined categories of work:

\begin{itemize}[noitemsep]
    \item Accounting \& Consulting
    \item Admin Support
    \item Data Science \& Analytics
    % \item Design \& Creative
    \item Engineering \& Architecture
    \item Sales \& Marketing
    \item Translation
    \item Web, Mobile \& Software Development
    \item Writing
\end{itemize}

To maintain domain alignment and evaluation consistency, experts are grouped into two broad specialization sets:

\begin{table}[h]
\centering
\caption{Expert set assignments by job category. Each set represents a group of Upwork professionals selected for their domain specialization and experience.}
\begin{tabular}{@{}p{4cm}p{10cm}@{}}
\toprule
\textbf{Set Name} & \textbf{Associated Job Categories} \\
\midrule
% \textbf{Design \& Creative} & Design \& Creative \\
\textbf{General} & Accounting \& Consulting; Admin Support; Sales \& Marketing; Translation; Writing \\
\textbf{Tech} & Web, Mobile \& Software Development; Engineering \& Architecture; Data Science \& Analytics \\
\bottomrule
\end{tabular}
\label{tab:expert_sets}
\end{table}

Within each set, assignments are balanced to ensure both coverage and cross-review. Experts receive domain-matched tasks to construct rubrics, assess deliverables, and provide per-criterion feedback. This organization supports consistent annotation standards while capturing the authentic variation in professional judgment across different fields.

By embedding qualified human professionals directly in the data-collection pipeline, UpBench maintains fidelity to real-world evaluation practices. This design ensures that the benchmark not only reflects genuine job structures and deliverables but also the evaluative perspectives of practitioners who perform such work in production environments.

\subsection{Filtering Jobs}

We begin from jobs that have been verified to be completed from Upwork. To ensure consistency and replicability, we restrict the dataset to fixed-price, single-milestone jobs with clearly defined requirements and verified client payments. Hourly projects are excluded to avoid ambiguity in scope and outcome quality.

\subsection{Additional Job Categorization}

After initial filtering, each qualified job was reviewed by domain-matched expert freelancers. They assigned two labels:
\begin{itemize}[noitemsep]
  \item task completeness: whether the job post contained sufficient context to be completed as written
  \item presence of personally identifiable information (PII): these labels support dataset quality control and analysis of data sensitivity.
\end{itemize}

For the benchmark project, freelancers did not access historical client–freelancer communications or prior deliverables. All assessments were based solely on job post content and any new materials created for this project, ensuring privacy compliance and a clear separation from prior marketplace activity.

This expert-led labeling adds practical human judgment while maintaining consistency and scalability. The framework also allows new tags or task-specific extensions without changing the overall pipeline. Qualified jobs were then verified by an additional expert, ensuring that each example reflects a real, completed project.

\subsection{Rubric Creation}

Each job is paired with a rubric—a structured set of acceptance criteria derived from the original job post, attachments, and deliverable context. Rubrics are designed to transform open-ended client instructions into measurable, verifiable requirements that support both automated and human evaluation.

Expert annotators construct between 5 and 20 rubric criteria per job, labeled according to their importance: critical, important, optional, or pitfall (criteria to avoid). This labeling scheme supports graded assessment rather than binary scoring and provides interpretable insights into agent performance. Rubrics are authored manually by domain experts, ensuring precision and domain alignment, and serve as both evaluation scaffolds and interpretive tools for analyzing agent errors.

\subsection{Evaluation of Job Deliverables}

AI model submissions are evaluated against the rubric criteria.  Human evaluators assess the same submissions, providing per-criterion feedback and justifications.  The per-criterion feedback, authored by expert freelancers, captures nuanced quality distinctions such as completeness, factual correctness, and alignment with client intent. These structured feedback traces are then stored alongside the job data to enable future research on reinforcement learning from human feedback (RLHF) and dynamic agent retraining.
\section{Dataset}

\subsection{Upwork Labor Marketplace Data}

\subsubsection{Job Context}

The UpBench comprises 322 real, economically verified jobs from Upwork, the world’s largest online labor marketplace where millions of freelancers complete professional contracts across domains such as software development, design, writing, analytics, and consulting. These jobs reflect genuine client demand, measurable financial transactions, and diverse task structures, focusing on simple, well-defined, and low complexity projects. By grounding benchmark tasks in this dynamic labor market, UpBench captures both the breadth of modern digital work and the heterogeneity of skills and evaluation standards required to succeed in it. The benchmark’s integration of rubrics, human feedback, and per-criterion grading provides a realistic and scalable framework for assessing how AI agents perform within real-world professional workflows, offering a bridge between experimental model evaluation and the real economics of online work.

Typical Upwork job posts combine a short task summary with explicit constraints that operationalize client intent into concrete work. Posts frequently specify (i) \emph{domain and audience} (e.g., German-language reviews for dating sites; Sinhalese transcription; sustainability-focused eCommerce titles), (ii) \emph{deliverables and format} (e.g., 3D renders, DOCX/PDF reports, spreadsheets, localized text), (iii) \emph{procedural instructions} (bullet lists, step-by-step checks, or acceptance rules such as “6 advantages and 3 disadvantages, $\leq$42 characters each” or “replace invalid emails before payment”), (iv) \emph{tools and platforms} (e.g., Excel, Unity, Mailchimp, Bootstrap, SQL/Python, or data sources like LinkedIn and YouTube), (v) \emph{scope and timelines} (word counts, quantity targets, or hard deadlines), and (vi) \emph{communication expectations} (attachments, milestone updates, preferred channels). The examples span content creation, translation, transcription, data collection, analytics, design, coding, and expert consulting. In practice, posts embed de facto acceptance criteria such as file types, quantitative thresholds, formatting standards, or quality gates (e.g., email validation), which UpBench extracts and normalizes into structured rubrics with per-criterion grading. This design captures the diversity and realism of digital work while making it systematically evaluable for agent assessment.

\subsubsection{Job Metadata}

The benchmark dataset contains contract-level data along with detailed job metadata capturing each project’s structure, scope, and categorization. Core fields include identifiers (\texttt{contract\_id}), temporal markers (\texttt{start\_ts}, \texttt{end\_ts}), and job descriptions (\texttt{job\_title}, \texttt{job\_description}). Payment information is represented by several fields (\texttt{job\_amount}, \texttt{contract\_amount}, \texttt{milestone\_amount}) depending on the contract type. Additional fields describe the job’s complexity and context, such as \texttt{expertise\_tier}, hierarchical categories (\texttt{category}, \texttt{subcategory}, \texttt{subsubcategory}), and meta indicators like the number of milestones or attachments.

% The dataset also includes ratings on automation potential (\texttt{automation\_desire\_rating}) and links to relevant O*Net tasks (\texttt{onet\_tasks}), providing a rich foundation for analyzing project characteristics and job-level heterogeneity.

\begin{figure*}[htbp]
  \centering

  % Row 1
  \begin{subfigure}[t]{0.47\textwidth}
    \centering
    \includegraphics[width=\linewidth]{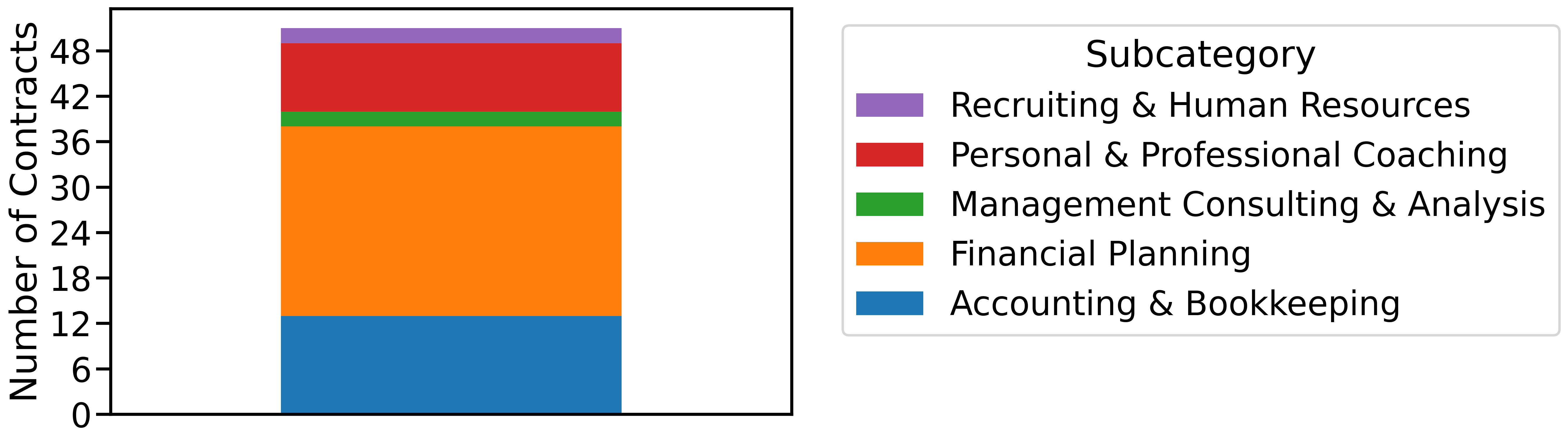}
    \caption{Accounting \& Consulting}
    \label{fig:cat_acc_consult}
  \end{subfigure}\hfill
  \begin{subfigure}[t]{0.47\textwidth}
    \centering
    \includegraphics[width=\linewidth]{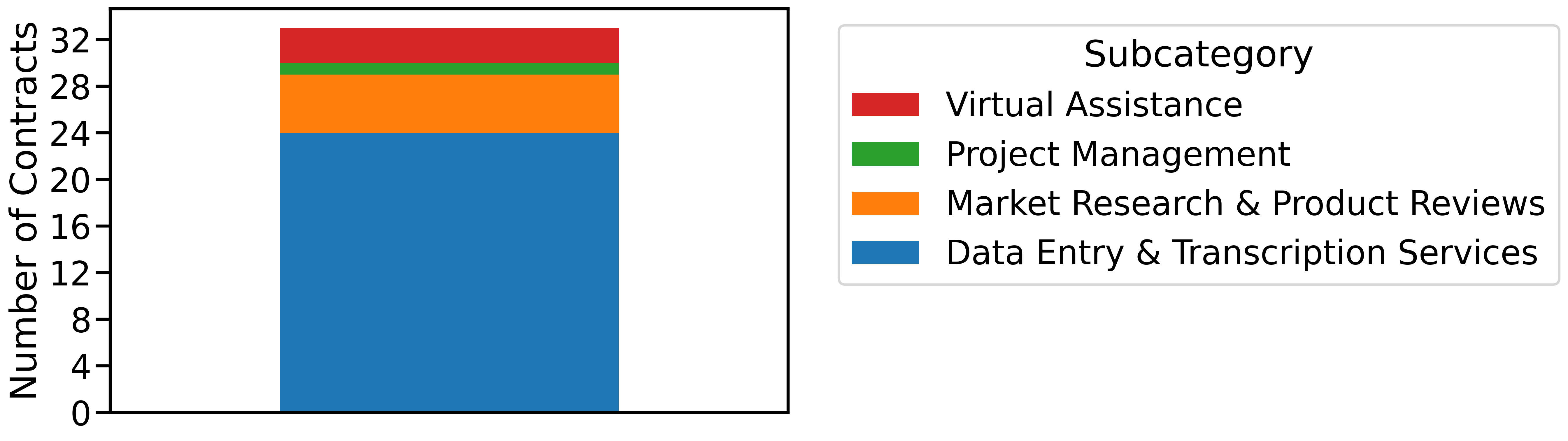}
    \caption{Admin Support}
    \label{fig:cat_admin}
  \end{subfigure}

  \vspace{10pt}

  % Row 2
  \begin{subfigure}[t]{0.47\textwidth}
    \centering
    \includegraphics[width=\linewidth]{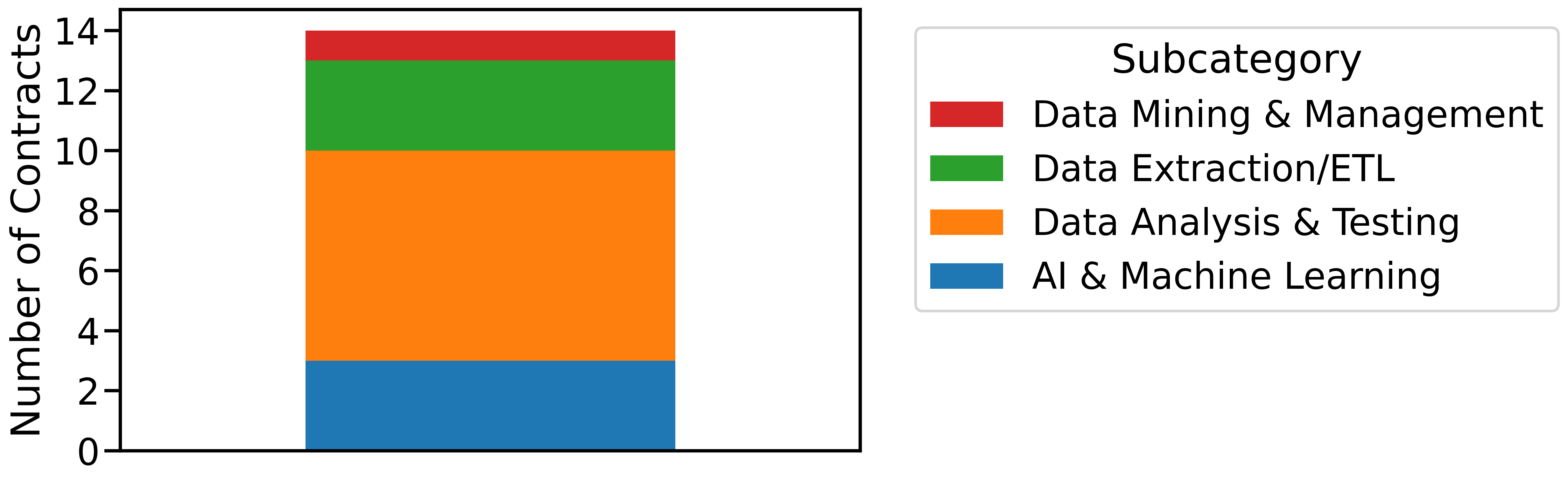}
    \caption{Data Science \& Analytics}
    \label{fig:cat_dsa}
  \end{subfigure}\hfill
  \begin{subfigure}[t]{0.47\textwidth}
    \centering
    \includegraphics[width=\linewidth]{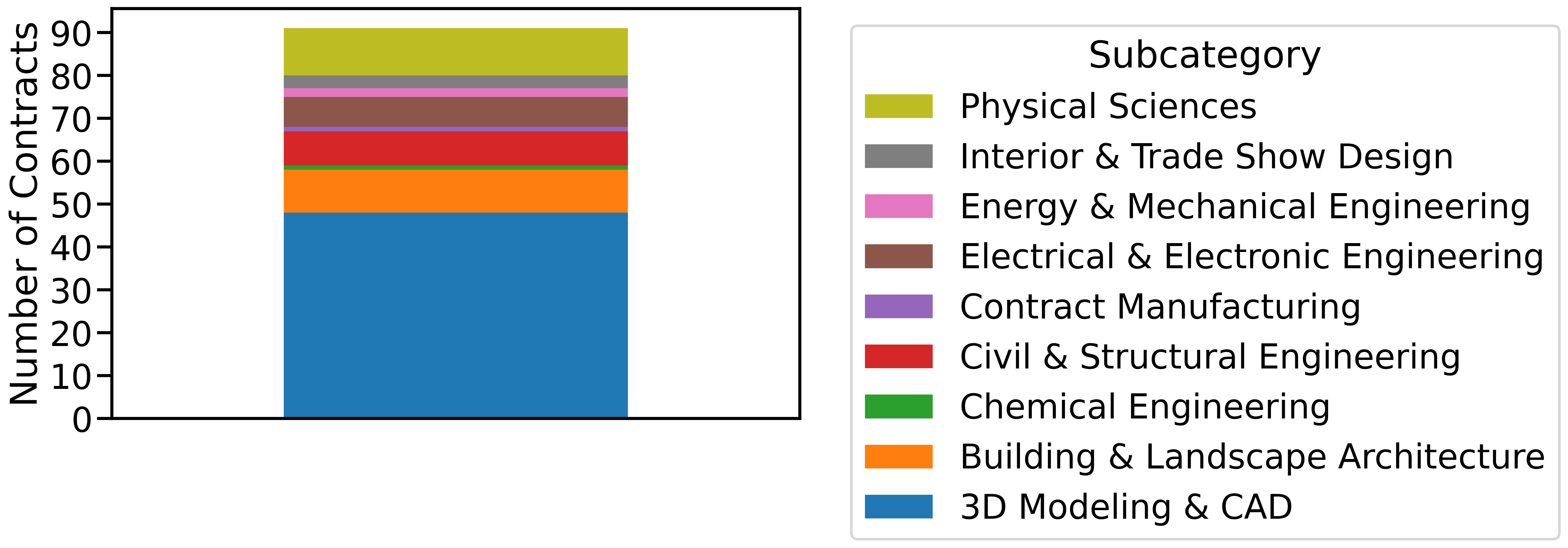}
    \caption{Engineering \& Architecture}
    \label{fig:cat_eng}
  \end{subfigure}

  \vspace{10pt}

  % Row 3
  \begin{subfigure}[t]{0.47\textwidth}
    \centering
    \includegraphics[width=\linewidth]{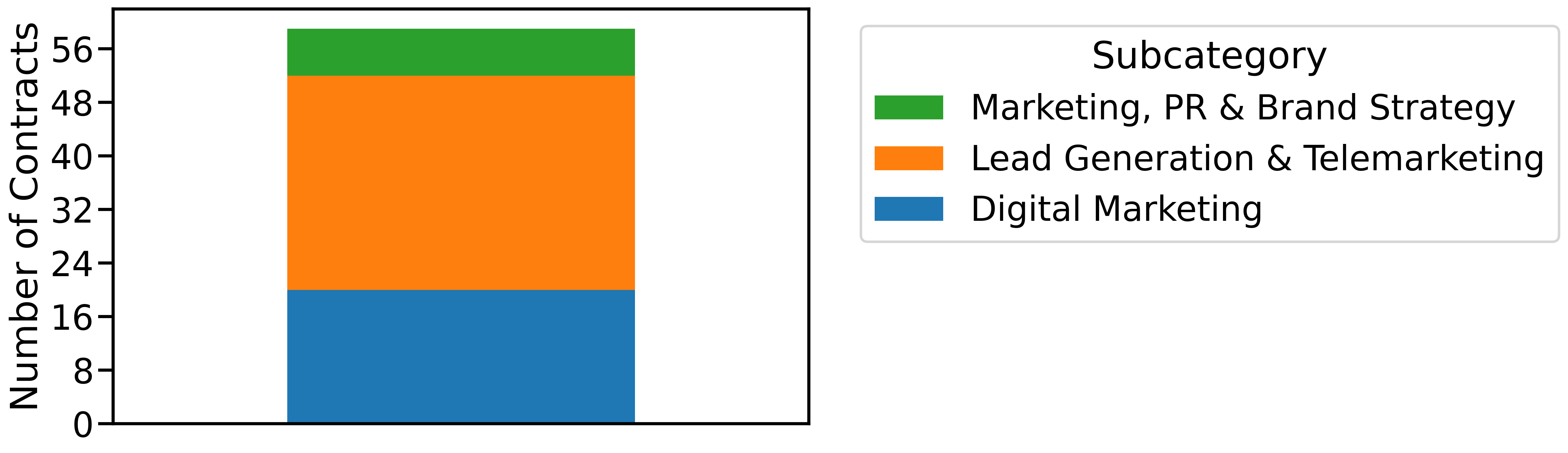}
    \caption{Sales \& Marketing}
    \label{fig:cat_sales}
  \end{subfigure}\hfill
  \begin{subfigure}[t]{0.47\textwidth}
    \centering
    \includegraphics[width=\linewidth]{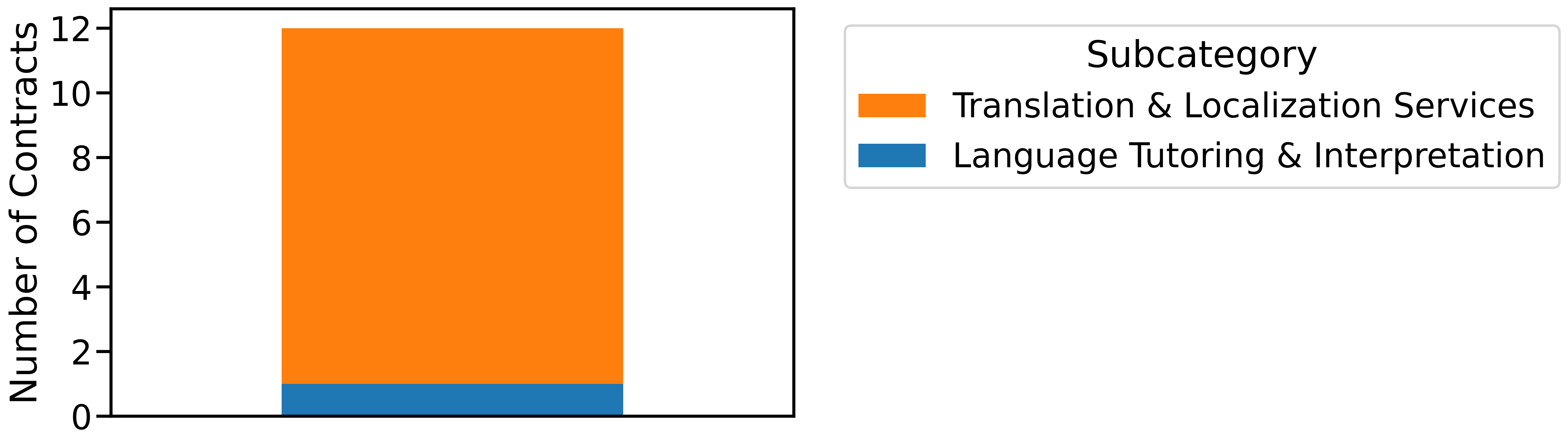}
    \caption{Translation}
    \label{fig:cat_translation}
  \end{subfigure}

  \vspace{10pt}

  % Row 4
  \begin{subfigure}[t]{0.47\textwidth}
    \centering
    \includegraphics[width=\linewidth]{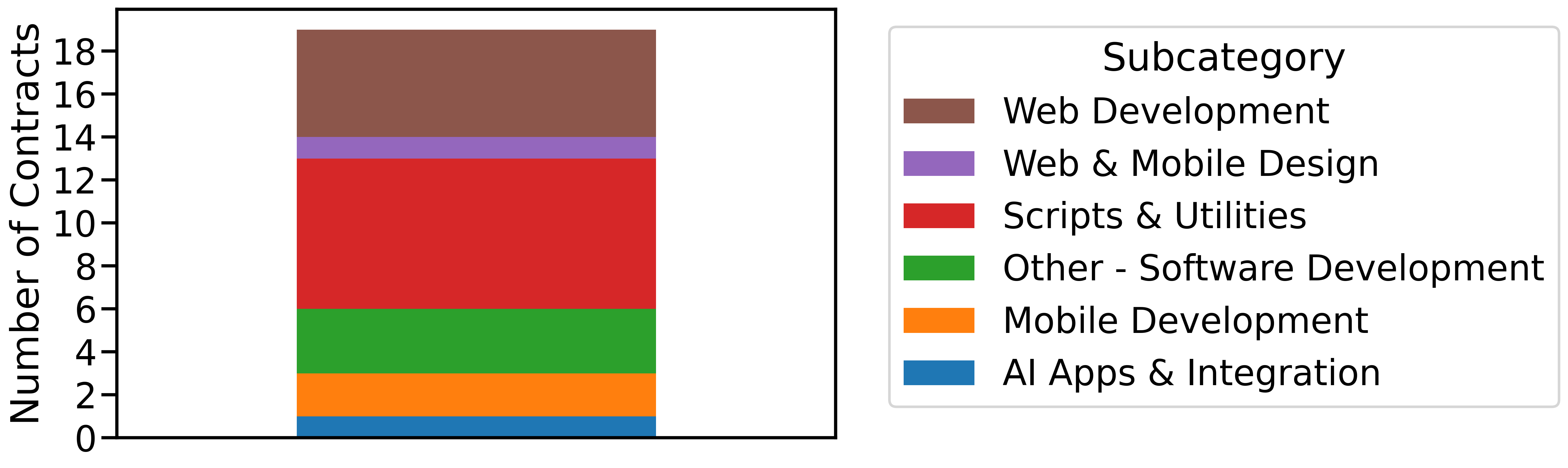}
    \caption{Web, Mobile \& Software Dev}
    \label{fig:cat_wmsd}
  \end{subfigure}\hfill
  \begin{subfigure}[t]{0.47\textwidth}
    \centering
    \includegraphics[width=\linewidth]{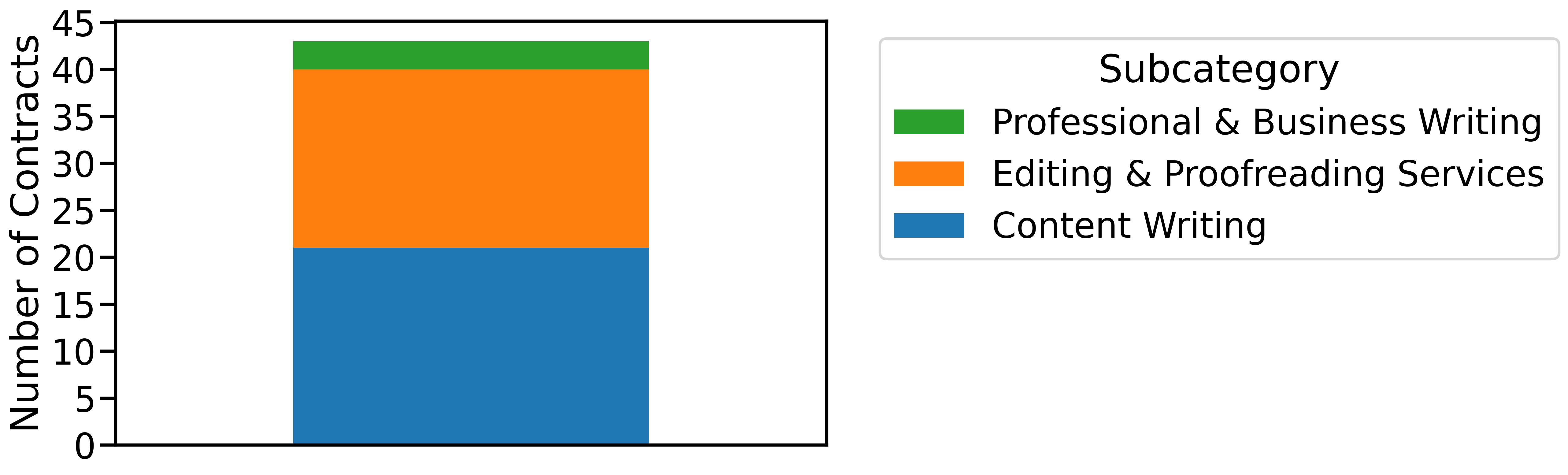}
    \caption{Writing}
    \label{fig:cat_writing}
  \end{subfigure}

  \caption{Distribution of subcategories within each of the nine UpBench job categories. Bars show the number of contracts observed per subcategory.}
  \label{fig:subcategory_grid}
\end{figure*}

Figure~\ref{fig:subcategory_grid} summarizes the distribution of contracts across subcategories within each of the eight job categories represented in UpBench.
Contract volumes are highest in Engineering \& Architecture, which spans multiple specialized fields—most prominently Civil/Structural Engineering and Electrical/Electronic Engineering—and in Sales \& Marketing, where Lead Generation \& Telemarketing dominates the category mix.
In Writing, most jobs fall under Content Writing and Editing \& Proofreading Services, while Admin Support is heavily concentrated in Data Entry \& Transcription Services.
Within Accounting \& Consulting, Financial Planning comprises the majority of contracts, and Translation remains overwhelmingly composed of Translation \& Localization Services.
Technical domains show broader dispersion: Web, Mobile \& Software Development includes substantial shares of Web Development and Scripts \& Utilities, with smaller but visible portions of AI Applications and Mobile Development.
Similarly, Data Science \& Analytics is led by Data Analysis \& Testing, supplemented by contributions from AI/ML and ETL subfields.

Overall, the subcategory spread reveals a heterogeneous and economically grounded task portfolio spanning operational, analytical, and engineering work.
This diversity underscores UpBench’s coverage of real-world digital labor—ranging from language-based deliverables and financial analysis to technical implementation and data-driven problem solving.

\clearpage
\subsubsection{Attachment Distribution}

\begin{figure}[h]
    \centering
    \includegraphics[width=0.7\linewidth]{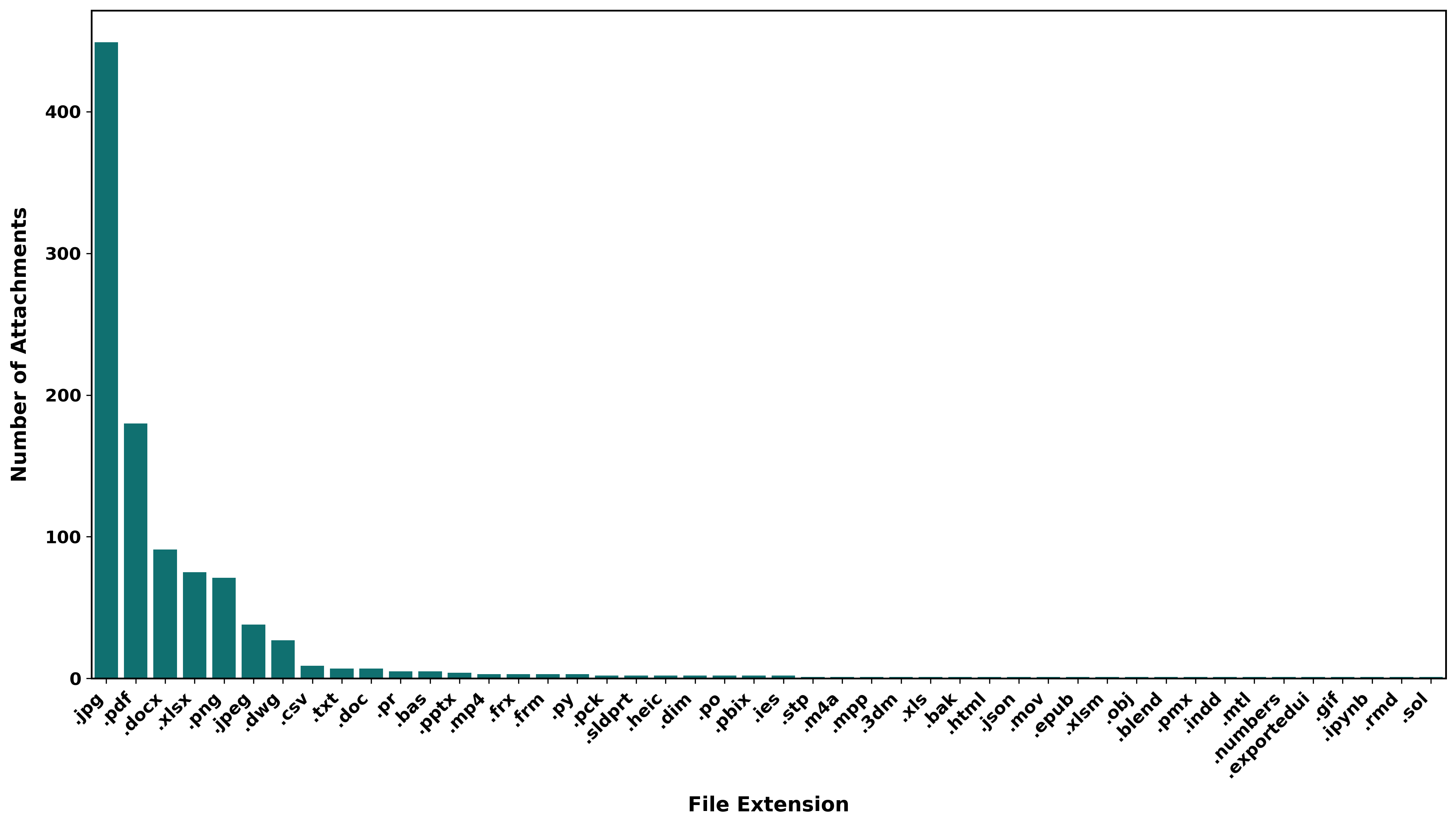}
    \caption{Extensions of Attachment Files}
    \label{fig:attachments}
\end{figure}

% \subsubsection{Deliverable File Distribution}
%\begin{figure}[h]
%    \centering
%    \includegraphics[width=0.7\linewidth]{figures/deliverables_types.png}
%    \caption{Extensions of Deliverables Files}
%    \label{fig:deliverables}
%\end{figure}

Figure~\ref{fig:attachments} shows the distribution of file types included as job attachments within the filtered UpBench dataset. The most common attachment formats are \texttt{JPG}, \texttt{PDF}, and \texttt{PNG}, followed by document and spreadsheet formats such as \texttt{DOCX} and \texttt{XLSX}. These patterns suggest that many job posts contain reference materials, design mockups, or textual briefs—consistent with professional service contexts like  analytics and documentation. The long tail of less frequent extensions (e.g., \texttt{.csv}, \texttt{.pptx}, \texttt{.mp4}) highlights the diversity of file modalities associated with real-world digital work, reflecting heterogeneous job requirements across Upwork categories.

%Figure~\ref{fig:deliverables} presents the distribution of file types submitted as final deliverables for completed jobs in the UpBench dataset. The majority of deliverables are \textbf{DOCX}, \textbf{PDF}, and \textbf{SVG} files, followed by structured data formats such as \textbf{XLSX}, \textbf{CSV}, and \textbf{JSON}. This composition indicates that most professional outputs are document-based or data-oriented, aligning with tasks in writing, design, analytics, and software development. The presence of code-related formats (e.g., \texttt{.py}, \texttt{.sql}, \texttt{.java}, \texttt{.js}) and media types (e.g., \texttt{.png}, \texttt{.ai}, \texttt{.mp4}) further underscores the benchmark’s cross-domain coverage. Compared to attachments, deliverables exhibit a higher concentration in text and code files, indicating a shift from input variety to structured, output-focused work artifacts.

% \subsubsection{Capturing Evolving Marketplace}

% FIGURE: TIME EVOLUTION OF POPULARITY OF DIFFERENT SUBSUBCATEGORIES OR KEYWORDS

\subsection{Rubrics}

Each job in the benchmark is accompanied by a rubric—a structured set of acceptance criteria derived directly from the job post, attachments, and deliverable context. These rubrics serve as a bridge between the client’s original, often ambiguous instructions and the objective evaluation required for reproducible benchmarking.

Rubric creation is performed by expert freelancers drawn from the relevant job category. Experts decompose the job requirements into 5–20 verifiable criteria, each designed to be specific, measurable, and aligned with the client’s stated goals. Criteria are labeled along four importance levels—critical, important, optional, and pitfall—following the labeling convention established in our previous work (\cite{terzolo2025upbench}).

\begin{itemize}[noitemsep]
    \item \textit{Critical} criteria must be satisfied for a deliverable to be considered acceptable.
    \item \textit{Important} criteria should be met and reflect strong alignment with the job objectives.
    \item \textit{Optional} criteria capture desirable but non-essential elements.
    \item \textit{Pitfall} criteria describe conditions to be explicitly avoided (e.g., factual errors, scope drift, or inappropriate content).
\end{itemize}

This schema allows fine-grained differentiation of quality across submissions while maintaining consistency across domains. By explicitly mapping the client’s goals to verifiable dimensions of quality, the rubric transforms open-ended professional work into a format suitable for both automated and human evaluation.  Full instructions given to freelancers for creating rubrics can be found in Appendix \ref{app:instructions}.

\subsection{Submission Feedback}

Each submission (whether produced by an AI agent or an AI+human collaboration) is evaluated against its corresponding rubric by a human freelancer.  Expert freelancers review the deliverables using the same rubrics. For each criterion, they provide binary scores (pass/fail) along with textual feedback describing observed strengths or deficiencies.

This hybrid setup mirrors the Evaluation Agent paradigm introduced in our previous environment framework (\cite{terzolo2025upbench}), where evaluators function as externalized reward models. The result is a set of granular, interpretable measurements that allow comparison across models, domains, and supervision modes (AI-only vs. human-in-the-loop).

Rubric-based evaluation ensures consistency and transparency while enabling future extensions, such as per-criterion reward shaping and dynamic feedback loops for iterative agent improvement.  Full instructions given to freelancers for evaluating deliverables can be found in Appendix \ref{app:instructions}.
\section{Experiments}

\subsection{Foundational LLM-powered Generalist AI Agents}

We evaluate three agent variants that share a common scaffold but differ in the underlying large language model (LLM) used to power them. Each agent operates within the same environment and employs identical prompting, tool use, and feedback protocols. The goal of this setup is to isolate differences in model capability while keeping the surrounding agentic logic and evaluation conditions constant.

All agents follow a simple plan–act–reflect loop inspired by ReAct-style reasoning (\cite{yao2023reactsynergizingreasoningacting}). Given a job description, attachments, and any relevant prior submissions, the agent first parses the task into sub-goals, generates a concise work plan, and then produces a deliverable aligned with the specified rubric. Each agent maintains access to lightweight tools for reading textual attachments and storing generated artifacts. No external retrieval, web access, or fine-tuning is performed during evaluation.

The three model-backed agents used in this study are:

\begin{itemize}[noitemsep]
    \item \textbf{Claude Sonnet 4 (Anthropic)}
    \item \textbf{Gemini 2.5 Pro (Google)}
    \item \textbf{GPT-5 (OpenAI)}
\end{itemize}

Each agent functions analogously to the Worker Agent described in our earlier multi-agent framework (\cite{terzolo2025upbench}): it receives qualified jobs curated through the UpBench pipeline, produces deliverables, and interacts with the corresponding rubrics and evaluators. The agents are intentionally minimal (no task-specific fine-tuning or external API orchestration), allowing the benchmark to measure intrinsic reasoning, adherence to instructions, and robustness across heterogeneous work domains.

\subsection{Key Metrics}
\subsubsection{Completion Rate}

We report two primary quantitative measures of performance: Completion Rate and Rubric Score.  We also report several auxiliary metrics that capture the economic and computational implications of agent behavior.

\begin{itemize}[noitemsep]
    \item \textbf{Completion Rate} measures the proportion of jobs for which an agent satisfies all required rubric criteria. A job is marked as completed only if every critical and important criterion passes. This strict metric represents full task completion under realistic client standards.
    \item \textbf{Rubric Score} captures partial task completion and provides a graded measure of quality. For each job, the Rubric Score is computed as the fraction of rubric criteria marked “pass”:
    % \begin{equation}
    %     \text{Rubric Score } = \frac{1}{N} \sum_{i=1}^N \frac{\text{criteria\_passed}_i}{\text{criteria\_total}_i}
    % \end{equation}
    Averaging across all jobs yields a dataset-level measure of how completely agents satisfy task requirements. The Rubric Score complements Completion Rate by quantifying near-misses and incremental improvements that may still represent usable outputs.
\end{itemize}

To better illustrate the different dimension of successul completion, we define the following system of evaluation and metrics, as seen in Table \ref{main:definitions}:

\begin{table}[h]
\centering
\caption{Definitions of Key Measures}
\label{main:definitions}
\begin{tabular}{lp{9cm}l}
\hline \hline
\textbf{Notations} & \textbf{Definition} & \\
\hline
\textbf{N} & Total number of jobs \textbf{attempted by AI} on the first try ($k = 1$) & \\
\textbf{A} & Number of jobs that \textbf{passed} on the first attempt ($k = 1$) & \\
\textbf{F = N - A} & Number of jobs that \textbf{failed} on the first attempt & \\
\textbf{M} & Number of \textbf{failed jobs} that received a HITL re-attempt ($k = 2$) & \\
\textbf{B} & Number of those $M$ HITL jobs that \textbf{passed} after re-attempt & \\
\hline
\textbf{$p_1 = \dfrac{A}{N}$} & Pass rate for \textbf{AI-only} (first attempt) & \\
$p_{H|F} = \dfrac{B}{M}$ & Conditional pass rate for \textbf{HITL among failed jobs} & \\
% $p_{AI|F} = \dfrac{B_{AI}}{M}$ & Conditional pass rate for \textbf{AI-only re-attempts among failed jobs} (control condition, if implemented) & \\
\textbf{$p_{\text{overall}} = \dfrac{A + B}{N}$} & Observed overall pass rate after adding HITL (lower bound $M < F$) & \\
% \textbf{$p_{\text{overall, true}} = \dfrac{A + B}{N - (F - M)}$} & Overall pass rate after adding HITL & \\
\hline
$R_{j,k} = \dfrac{\#\text{criteria passed}_{j,k}}{\#\text{criteria total}_{j,k}} \in [0,1]$  & Rubric pass score for each job $j$ and attempt $k$ & \\
\textbf{$\bar{R}_1 = \dfrac{1}{|N|}\sum_{j \in N} R_{j,1}$} & Average rubric score on first attempt ($k=1$) & \\
$\bar{R}_{2 \mid M} =\dfrac{1}{|M|}\sum_{j \in M} R_{j,2}$ & Conditional rubric pass score for HITL among failed jobs & \\
$\bar{R}_{\text{overall}} \;=\; \dfrac{1}{|N|}\sum_{j \in N} R_{j,\text{final}}$ & where 
$
R_{j,\text{final}} \;=\;
\begin{cases}
R_{j,1}, & \text{if } j \in A \text{ (passed at } k=1)\\[4pt]
R_{j,2}, & \text{if } j \in M \text{ (HITL attempt at } k=2)\\[4pt]
R_{j,1}, & \text{if } j \in F \setminus M \text{ (failed, no re-attempt)}
\end{cases}
$ & \\
\hline \hline
\end{tabular}
\end{table}

\newpage

\subsubsection{Economic Efficiency and Value}

Beyond these two core indicators, we also report secondary performance metrics that provide additional interpretive context:

\begin{itemize}[noitemsep]
    \item \textbf{Economic Value Captured:} By linking each job to its historical payout, we can estimate the real-world monetary value of successfully completed tasks. Summing payout values over successful submissions yields an approximate measure of ``earnings potential'' for each agent, a proxy for practical economic utility in real labor-market conditions.
    \item \textbf{Runtime Efficiency and Cost:} We compare both median completion time and token-level cost across models to assess the trade-off between performance and computational efficiency. These measures capture the time and resource requirements of each model configuration, supporting a holistic view of performance that integrates quality, speed, and cost.
\end{itemize}

Together, these metrics provide a comprehensive evaluation of model effectiveness.

\subsection{Re-attempts Through Human Feedback}

In addition to fully autonomous evaluation, we include a human-in-the-loop (HITL) condition to assess how effectively agents incorporate feedback from expert evaluators. This setting mirrors real-world workflows where humans review, critique, and guide AI-generated work.

The HITL process proceeds in four steps, following the protocol from our prior studies (\cite{terzolo2025upbench}):

\begin{algorithm}[H]\label{alg:hitl}
    \SetAlgoLined
    \KwData{job $\mathcal{J}$, rubric $\mathcal{R}$, agent model $M$, human evaluator $H$, max iterations $K$, cost budget $B$}
    \KwResult{Final submission $S^{\star}$, history $\mathcal{H}$ of attempts and feedback}
    initialize iteration counter $t \gets 0$, cumulative cost $b \gets 0$\;
    initialize submission $S_0 \gets M(\mathcal{J}, \mathcal{R})$\;
    evaluate submission: $(\text{grade}_0, F_0, p_0) \gets H(S_0, \mathcal{R})$\;
    record $(S_0, \text{grade}_0, F_0, p_0)$ in history $\mathcal{H}$\;
    $b \gets b + \textsc{Cost}(\text{grade}_0)$\;
    \While{$p_t = \textsc{Fail}$ \textbf{and} $t < K$ \textbf{and} $b < B$}{
        \tcc{Agent revises submission using human feedback}
        $t \gets t + 1$\;
        $S_t \gets M(\mathcal{J}, \mathcal{R}, S_{t-1}, F_{t-1})$\;
        evaluate new submission: $(\text{grade}_t, F_t, p_t) \gets H(S_t, \mathcal{R})$\;
        record $(S_t, \text{grade}_t, F_t, p_t)$ in $\mathcal{H}$\;
        $b \gets b + \textsc{Cost}(\text{grade}_t)$\;
        \If{$\textsc{RubricScore}(\text{grade}_t) \le \textsc{RubricScore}(\text{grade}_{t-1})$}{
            \textbf{break} \tcc*[r]{Optional early stop if no improvement}
        }
    }
    \caption{Human-in-the-Loop (HITL) Iterative Refinement}
\end{algorithm}

We denote each attempt by the AI Agent with the variable $k$.  For example, the first AI-only attempt would be $k=1$ and the following feedback-fueled re-attempt would be $k=2$.  This is in line with the pass@k formulation from \cite{chen2021evaluating}.

This setup allows direct measurement of an agent’s responsiveness to human feedback and its capacity for self-correction. Comparing the Completion Rate and Rubric Score between AI-only and HITL conditions quantifies the marginal benefit of human guidance.

Beyond performance improvement, the HITL condition provides structured training data for future agent learning loops—linking per-criterion feedback to behavioral adjustments. It also serves as a diagnostic tool for understanding failure recovery, instruction following, and adaptation in open-ended professional tasks.

%\clearpage
\subsection{Results}

\begin{table}[htbp]
\centering
\caption{Performance Summary by Agent}
\label{main:performance}
\begin{tabular}{lrrrrr}
\hline
\textbf{AI Agent} & \textbf{P1} & \textbf{P Overall} & \textbf{Abs Lift} & \textbf{Relative Lift} & \textbf{Rescue Rate} \\
\hline
Claude Sonnet 4 & 0.398 & 0.512 & 0.115 & 0.289 & 0.233 \\
Gemini 2.5 Pro & 0.199 & 0.323 & 0.124 & 0.625 & 0.180 \\
GPT-5 & 0.196 & 0.335 & 0.140 & 0.714 & 0.190 \\
\hline
\end{tabular}
\end{table}

\begin{figure}[h]
    \centering
    \includegraphics[width=0.7\linewidth]{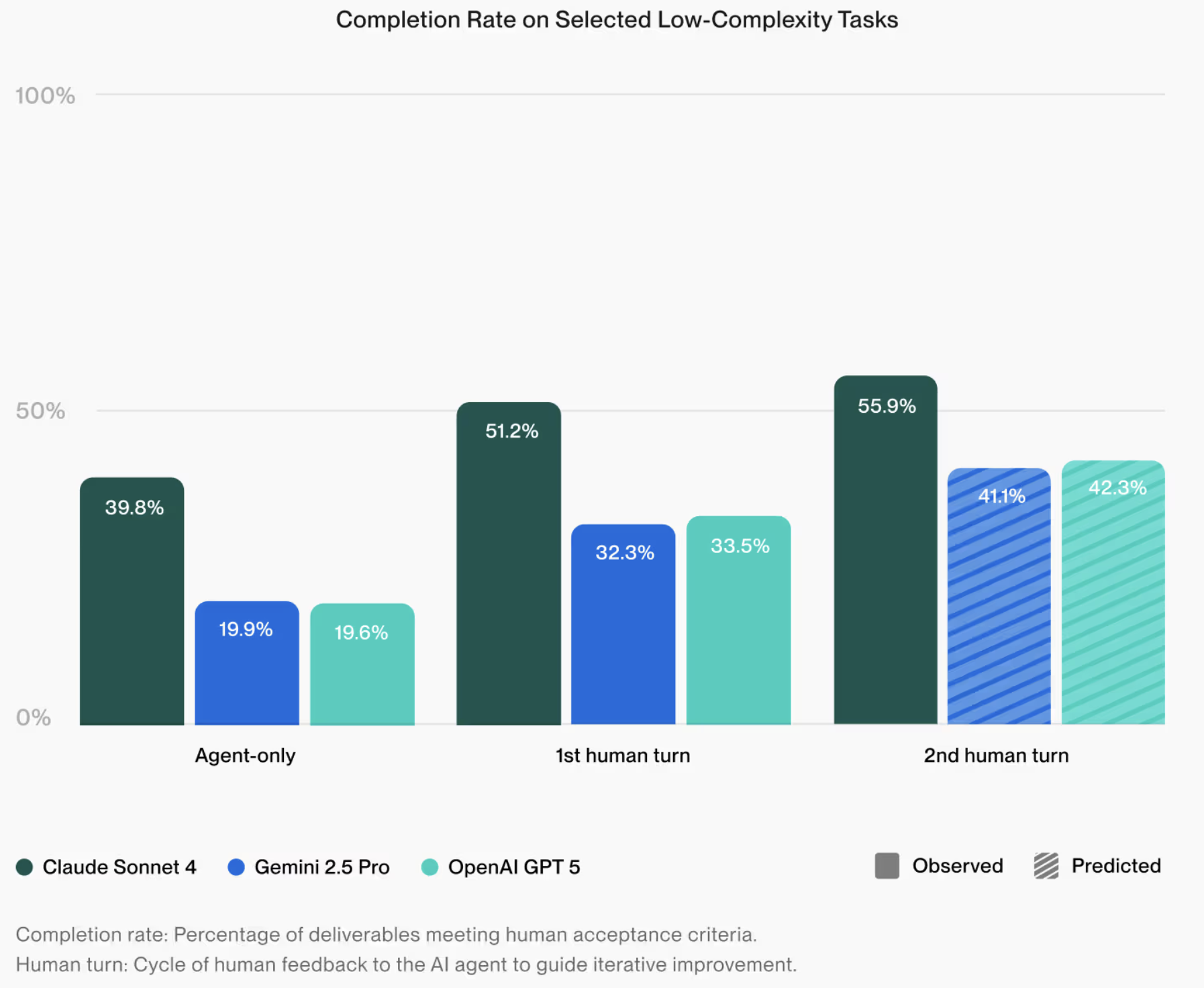}
    \caption{Project Completion    Rate: AI Only vs. HITL}
    \label{fig:successrate}
\end{figure}

Table \ref{main:performance} and Figure \ref{fig:successrate} reports performance metrics for three AI agents evaluated under both AI-only (k=1) and Human-in-the-Loop (HITL, k=2) conditions. We additionally show the k=3 HITL loop for our claude model as well as model-based predictions for the k=3 value of the other models (see Appendix \ref{app:modelling} for details).  The P$_1$ column represents the initial success rate on first attempt, while P Overall captures the overall success rate after incorporating human feedback and allowing re-attempts. The Absolute Lift ($\Delta p$) measures the gain in success rate from adding HITL, whereas the Relative Lift expresses this improvement as a percentage relative to the baseline. The Rescue Rate quantifies the share of initially failed jobs that were successfully “rescued” through human-guided refinement.

Across all three agents, the integration of human feedback leads to substantial performance improvements, exceeding those observed in the initial release. Absolute lifts in success rate now range from 0.115 to 0.140 (11–14 percentage points), corresponding to relative gains between 29\% and 71\% over the AI-only baseline. The Rescue Rate, which measures the share of initially failed jobs successfully recovered through human-guided refinement, rises to between 18\% and 23\%—indicating that nearly one in five failed tasks can be salvaged through a single HITL iteration. These stronger gains demonstrate the growing effectiveness and generality of the human-in-the-loop process: even limited expert feedback produces measurable, domain-wide improvements in task success and reliability across heterogeneous model architectures.

% \begin{figure}[h]
%     \centering
%     \includegraphics[width=0.8\linewidth]{figures/all_performance_metrics_by_agent.png}
%     \caption{Caption}
%     \label{fig:allmetrics}
% \end{figure}

\begin{figure}[h]
    % \begin{subfigure}[b]{0.48\textwidth}
        \centering
    \includegraphics[width=0.48\textwidth]{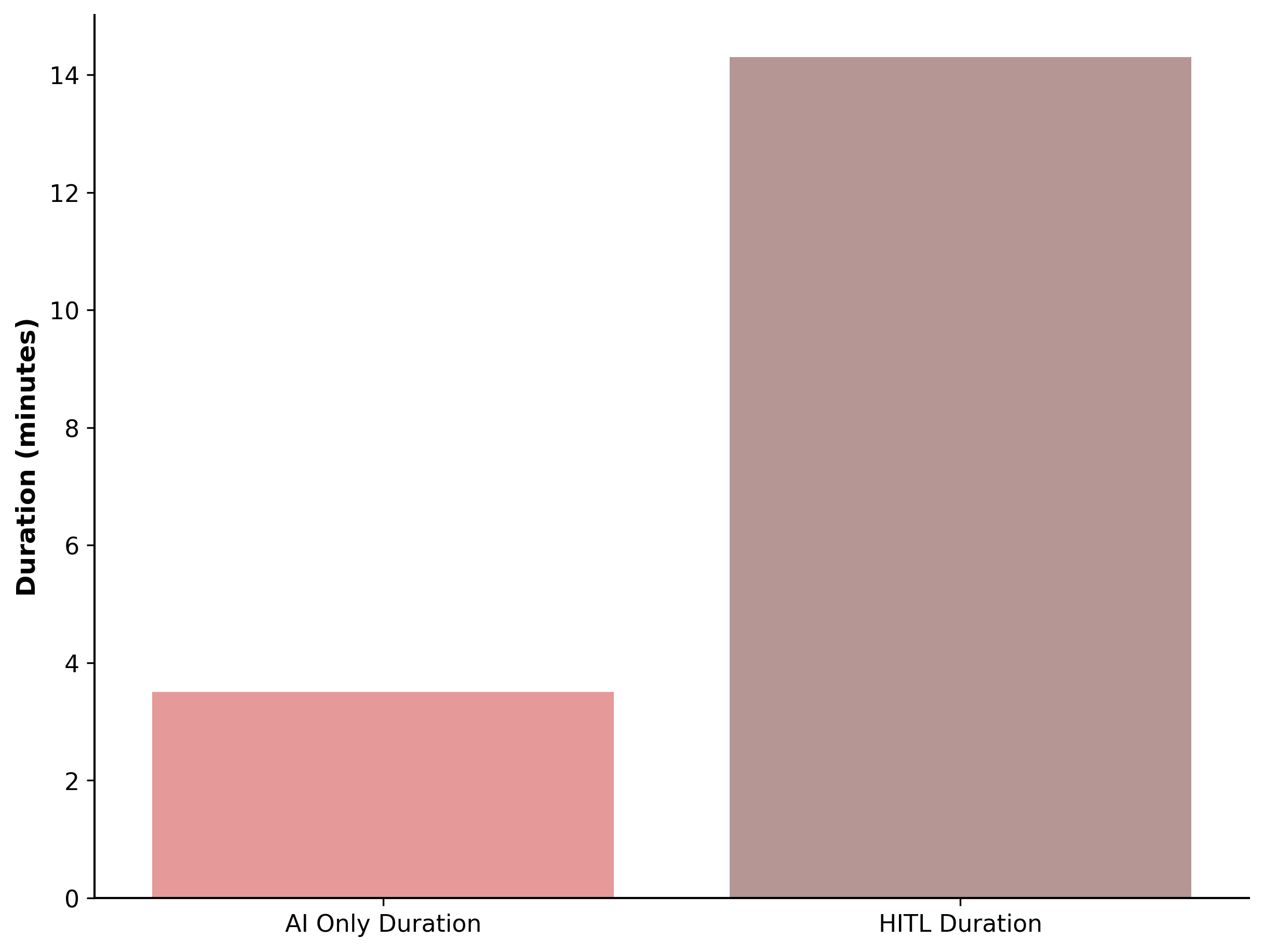}
        \label{fig:compare:duration}
    % \end{subfigure}
    \caption{Duration Comparison between AI Only and HITL}
    \label{fig:compare:both}
\end{figure}

Figure \ref{fig:compare:both} compares the duration between AI-only and Human-in-the-Loop (HITL). While HITL incurs additional time, it also unlocks measurable economic value, as seen in Table \ref{main:performance}, supporting its role in higher-stakes or complex jobs where accuracy justifies the investment.

\begin{figure}[h]
    \centering
    \begin{subfigure}[b]{0.8\textwidth}
        \centering
        \includegraphics[width=\linewidth]{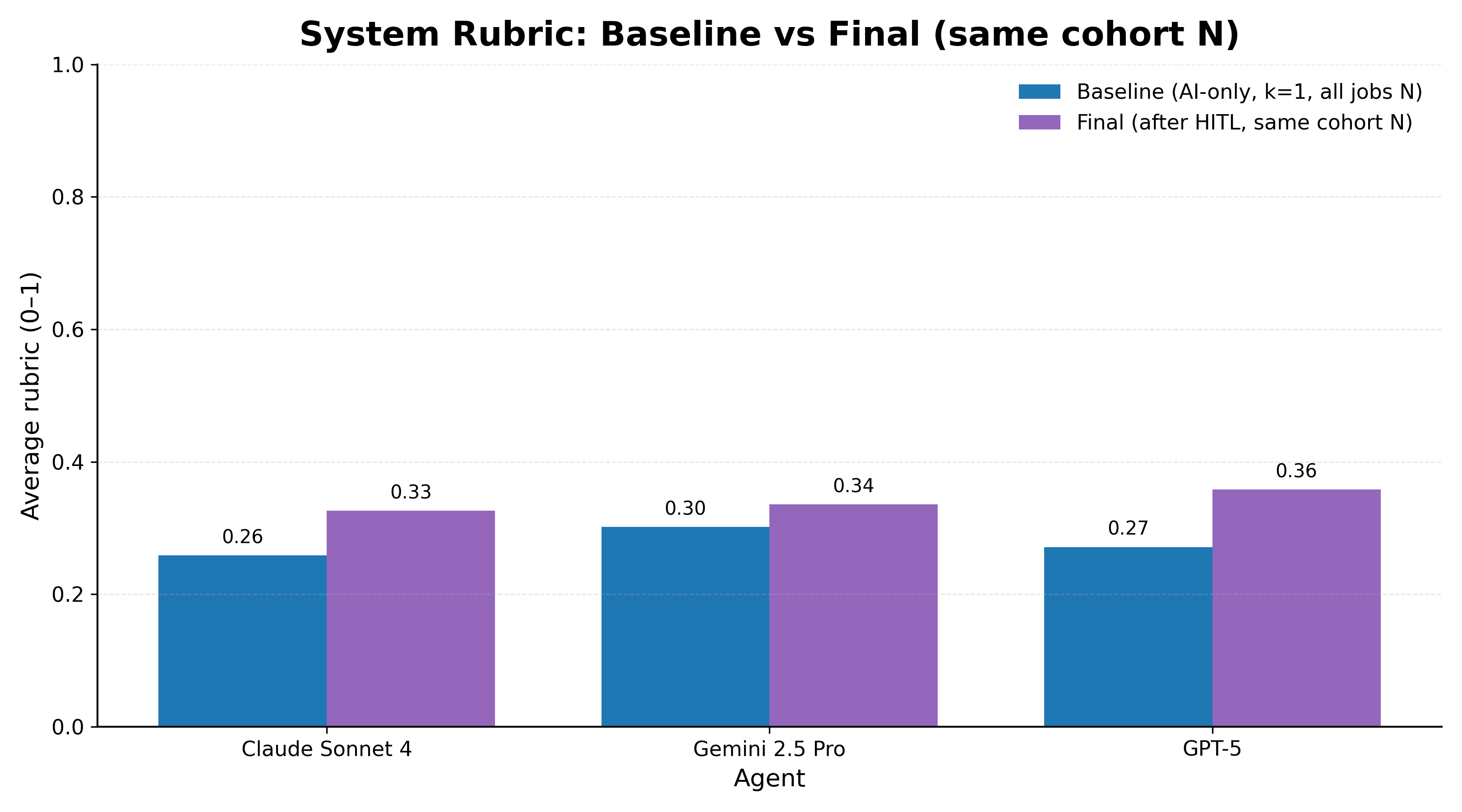}
        \label{fig:compare:rubricabs}
    \end{subfigure}
    
    \vspace{0.5em} % Adjust vertical spacing between figures

    \begin{subfigure}[b]{0.8\textwidth}
        \centering
        \includegraphics[width=\linewidth]{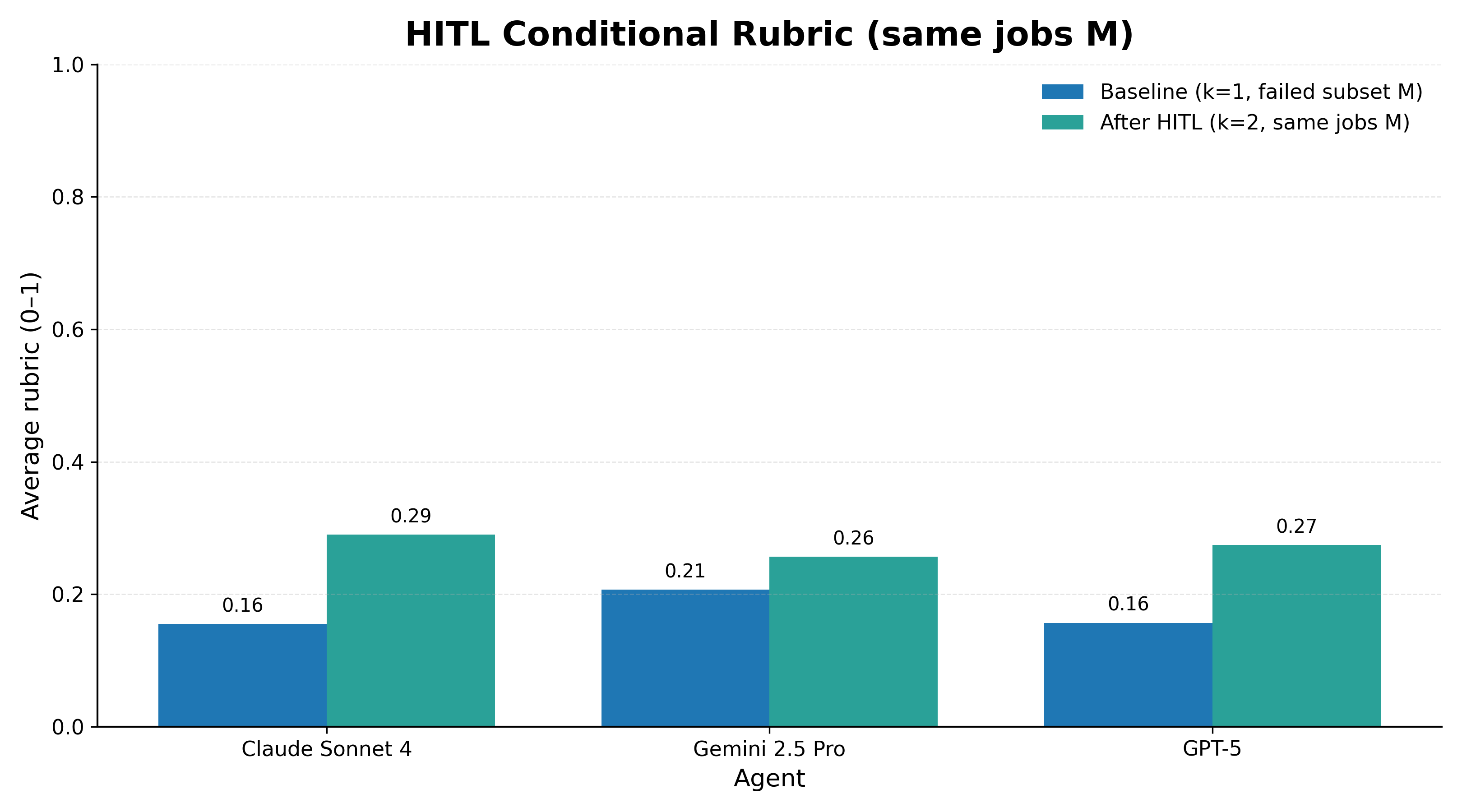}
        \label{fig:compare:rubricrelative}
    \end{subfigure}
    
    \caption{Overall comparison of absolute and relative rubric lift.}
    \label{fig:compare:rubricboth}
\end{figure}

Figure \ref{fig:compare:rubricboth} compares rubric-based performance across the three AI agents, showing average rubric scores before and after the HITL process over the same set of jobs. Across all models, the final rubric scores (purple) consistently exceed baseline AI-only scores (blue), demonstrating that human feedback and re-attempts yield substantial quality improvements. The average absolute rubric lift now ranges from 0.06 to 0.09 on a 0–1 scale—more than double the previous release—representing clear, consistent gains in overall task quality.

The second panel isolates the subset of initially failed jobs that underwent human-in-the-loop re-attempts. Here, each agent exhibits marked improvement after HITL intervention, with relative rubric gains of roughly 40–70\%. This conditional view captures the direct effect of human feedback, showing that evaluators help “rescue” a large share of previously unsuccessful attempts. Together, these results reinforce the value of iterative, feedback-driven refinement in the UpBench framework, where even limited human guidance substantially elevates agent performance quality across heterogeneous work domains.

\begin{figure}[h]
   \centering
   \includegraphics[width=0.8\linewidth]{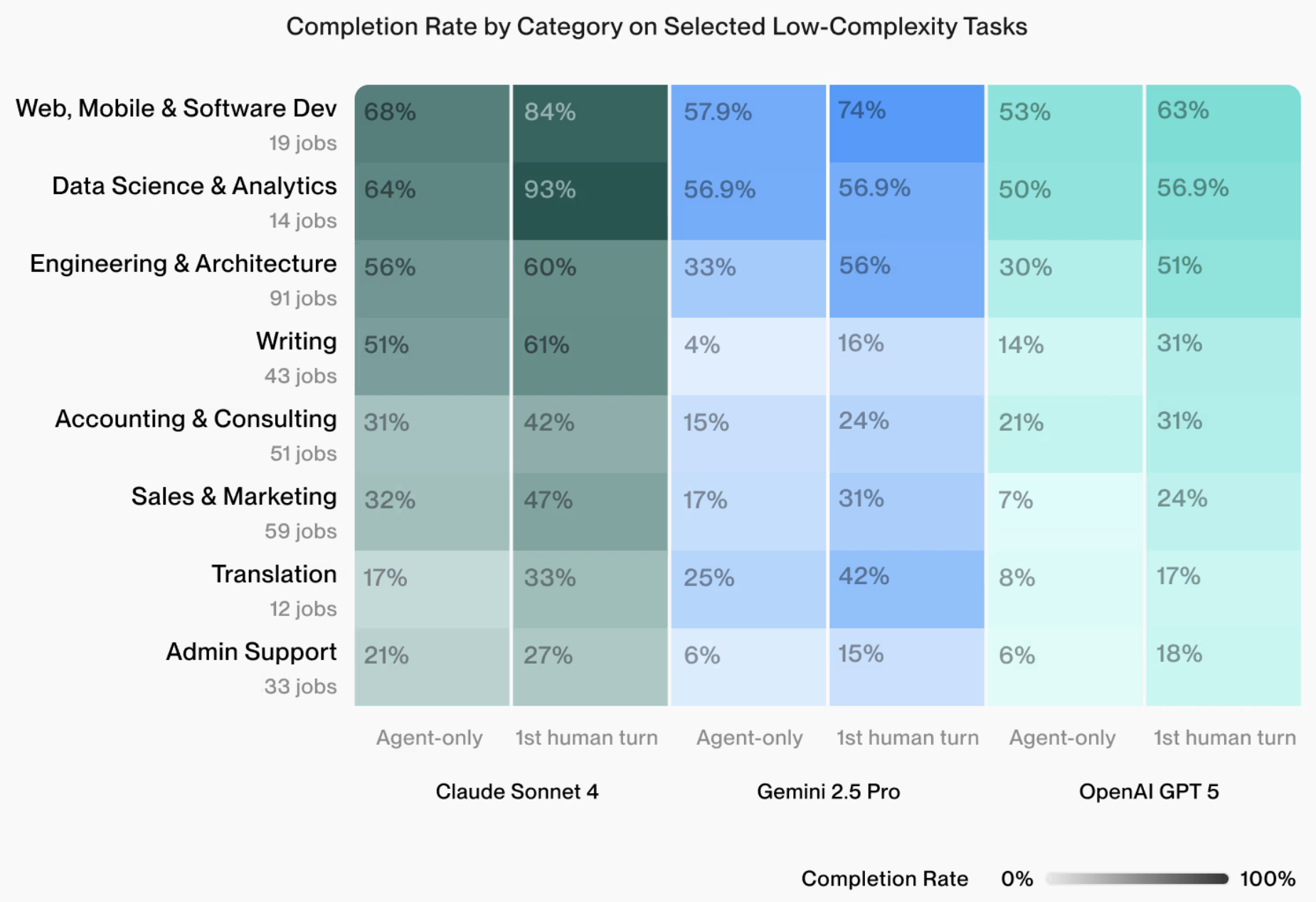}
   \caption{Pixel heatmap of completion rate change ($\Delta$SR) by job categories for three agents.}
   \label{fig:heatmap}
\end{figure}

The heatmap Figure \ref{fig:heatmap} summarizes how top-$k$ settings affect agent performance across fine-grained job types. Each row corresponds to a distinct category, while the six data columns show $\Delta$SR for Claude Sonnet 4.0, Gemini 2.5 Pro, and GPT-5 at $k{=}1$ and $k{=}2$ (left to right within each agent block). Cell color encodes success-rate change. The categories are ordered by average completion rates with higher completion rates at the top.

\subsubsection{Value Comparison}

Based on the previous results, we model the expected economic value of completing a task using one of several approaches—AI-only, Human-in-the-loop (HITL), or Human-only—based on their success probabilities, costs, and the value of a successful outcome.

For each approach $i$, the expected net value is given by:

\begin{equation}
E(V_i) = s_i \cdot V - c_i
\end{equation}
where:
\begin{itemize}
    \item $s_i$ = success rate (probability of success)
    \item $V$ = economic value of the task if successful
    \item $c_i$ = cost (time, money, or resources) of that approach
\end{itemize}

This equation represents the expected net payoff to the buyer or organization for using method $i$.

The breakeven task value between two approaches, say method 1 and method 2, occurs when their expected net values are equal:

$$
s_{H} \cdot V^* - c_H = s_L \cdot V^* - c_L
$$

Solving for $V^*$ gives:

$$
V^* = \frac{c_H - c_L}{s_H - s_L}
$$

If $V > V^*$, the higher-success, higher-cost approach (e.g., HITL) yields greater expected net value.
If $V < V^*$, the lower-cost, lower-success approach (e.g., AI-only) is more efficient.

\textbf{Decision Regions}
Depending on the breakeven thresholds, the optimal approach can be described by:
$$
\text{Choose AI-only if } \quad V < V_{\text{AI$\leftrightarrow$HITL}}
$$
$$
\text{Choose HITL if } \quad V_{\text{AI$\leftrightarrow$HITL}} \le V < V_{\text{HITL$\leftrightarrow$Human}}
$$
$$
\text{Choose Human-only if } \quad V \ge V_{\text{HITL$\leftrightarrow$Human}}
$$

This framework allows direct comparison of cost–performance tradeoffs and helps quantify the economic value of human–AI collaboration.

\begin{figure}[ht]
    \centering
    \includegraphics[width=0.8\linewidth]{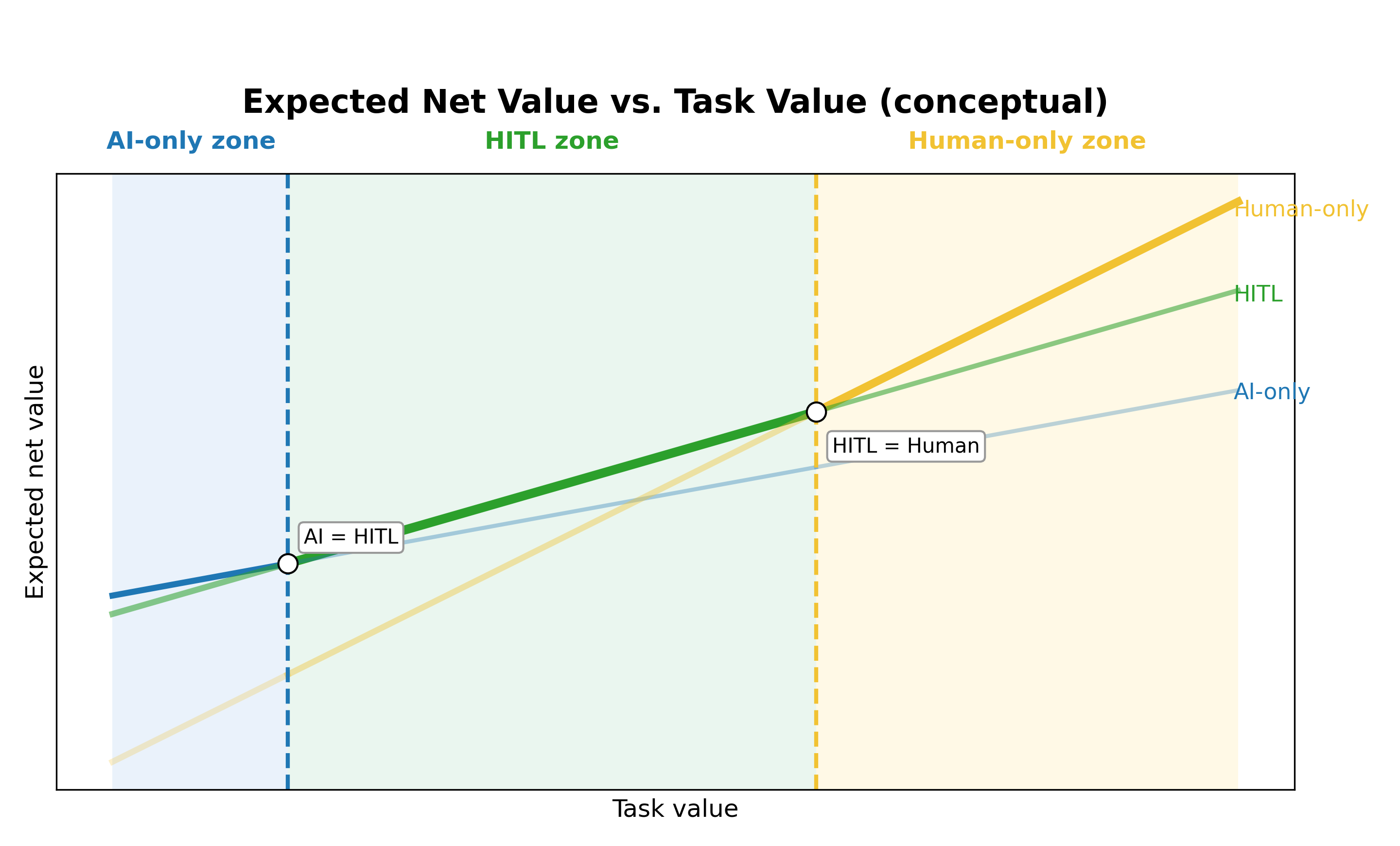}
    \caption{Expected Net Value by Task Value Across AI, HITL, and Human Approaches}
    \label{fig:value}
\end{figure}

Figure \ref{fig:value} illustrates the relative economic value of three approaches—AI-only, Human-in-the-Loop (HITL), and Human-only—across tasks of increasing value.
At low task values, AI-only solutions yield the highest expected net value because their cost is minimal and the marginal benefit of greater accuracy is small.
As task value rises, HITL systems overtake AI by offering higher success rates that justify their added human cost.
For very high-value or high-risk tasks, full human execution becomes optimal despite its expense, as the cost of failure outweighs automation gains.

The shaded regions correspond to these regimes: the blue AI-only zone for low-value work, the green HITL zone for mid-value tasks where collaborative feedback adds the most value, and the gold Human-only zone for high-value cases requiring full reliability.
The dashed vertical lines mark the conceptual thresholds where each method’s expected net value intersects—AI = HITL on the left and HITL = Human on the right.\footnote{These boundaries are based on simulation of observed costs and success rates in our sample and should be interpreted as approximate, data-driven thresholds rather than fixed universal values; they will shift with model quality, task type, and human labor cost.}

\subsection{LLM-aided Cluster Analysis}

To better understand patterns in rubric design and agent performance, we conduct an LLM-Aided Cluster Analysis over all rubric criteria collected in the UpBench dataset.  Each rubric criterion represents a short, natural-language statement describing a specific deliverable requirement (e.g., ``the report must include data visualizations'' or ``the code must run without external dependencies'').  Analyzing these at scale provides insight into which classes of requirements are most prevalent and where AI systems tend to underperform.

We begin by embedding all rubric criteria using OpenAI’s text-embedding-3-large model, which produces 3,072-dimensional dense representations that capture semantic similarity between textual instructions.  These embeddings are normalized and then subjected to hierarchical agglomerative clustering with cosine distance as the similarity metric.  The clustering process identifies high-level groupings of criteria that share conceptual similarity (e.g., formatting rules, factual correctness, reasoning requirements).  We select the top 20 clusters using distance-based cut thresholds to balance interpretability and coverage across the dataset.

To interpret each discovered cluster, we employ GPT-5 as an analytical assistant.  For every cluster, GPT-5 receives the list of rubric criteria and generates a concise textual summary describing the unifying concept and representative examples.  This process provides human-readable labels for otherwise latent semantic clusters, producing interpretable themes such as data accuracy, structure and clarity, or adherence to instructions.  The LLM-assisted summaries serve both as descriptive metadata and as anchors for subsequent quantitative analysis.

After clustering, we align each rubric criterion with recorded agent evaluation outcomes.  This enables computation of failure frequency per cluster—how often AI agents fail specific categories of criteria relative to others.

\begin{figure}[htbp]
    \centering
    
    % Left subfigure
    \begin{minipage}[t]{0.45\textwidth}
        \centering
        \textbf{(a) Common Criteria Clusters} \\[4pt]
        \begin{itemize}[noitemsep]
            \item Deliverable format and content requirements - 18.45\%
            \item Spreadsheet columns and row requirements - 7.53\%
            \item Language Quality and Formatting Standards - 6.77\%
            \item Automated structured valuation workbook - 5.13\%
            \item Financial report structure and data updates - 3.38\%
            \item Word count \& formatting constraints - 2.29\%
            \item Translation accuracy, completeness, and formatting - 1.97\%
            \item Survey response completeness \& formatting - 1.97\%
            \item Tailored Resume \& Cover Letter Requirements - 1.86\%
            \item Spreadsheet template adherence \& completeness - 1.86\%
        \end{itemize}
    \end{minipage}
    \hfill
    % Right subfigure
    \begin{minipage}[t]{0.45\textwidth}
        \centering
        \textbf{(b) Failed Criteria's Clusters} \\[4pt]
        \begin{itemize}[noitemsep]
            \item Deliverable format and content requirements - 15.49\%
            \item Spreadsheet columns and row requirements - 9.37\%
            \item Financial report structure and data updates - 7.48\%
            \item Automated structured valuation workbook - 6.55\%
            \item Language Quality and Formatting Standards - 5.58\%
            \item UK area K-means maps \& outputs - 3.52\%
            \item Task checklist table with six fields - 2.82\%
            \item Translation accuracy, completeness, and formatting - 2.44\%
            \item Word count \& formatting constraints - 2.17\%
            \item Event Marketing and Ad Creative Optimization - 2.11\%
        \end{itemize}
    \end{minipage}
    
    \caption{Criteria Clusters.}
    \label{fig:criteriaClusters}
\end{figure}
\section{Limitations}
While this study offers a first large-scale evaluation of human-in-the-loop (HITL) performance in agentic AI systems, several limitations should be noted. Below we list some of the strongest limitations for this study.

\textbf{Rubrics $\neq$ client acceptance.} Our acceptance rubrics emphasize objective, quick-to-verify checks, which do not fully capture qualitative factors (e.g., taste, polish, communication, iteration) that influence real client acceptance.

\textbf{Minimal agent scaffolding.} The evaluated agents were minimal wrappers over foundational LLMs (simple plan--act--reflect; no retrieval, web tools, or fine-tuning), so results are not representative of tool-rich, state-of-the-art agent systems.  This study’s scope is limited to a few models, domains, and up to two re-attempts.

\textbf{Restricted contract types.} The first release scoped jobs to \emph{fixed-price, single-milestone} contracts, omitting hourly and multi-milestone workflows that add timeline, coordination, and negotiation complexity. Additionally, our sampling was designed for feasibility rather than to mirror the marketplace-wide job distribution, which can induce category skews and limit generalization.  Finally, a corpus of \textbf{322} jobs cannot capture the full diversity of Upwork tasks, deliverable modalities, and difficulty levels.  We hope to expand the benchmark to capture more complex jobs that better represent all of the jobs on the Upwork Marketplace.

\textbf{Inter-rater variability unmeasured.} While emphasizing consistency and replicability, the study does not yet quantify inter-rater agreement or report error margins. Future work should include multiple independent graders, estimate inter-rater reliability, and present confidence intervals to capture evaluation uncertainty.

\textbf{HITL lift lacks AI-only ablation.} We did not run matched \emph{AI-only} second attempts on the same failed jobs, so the benefit of human feedback cannot be isolated from a ``second-chance'' effect.

\textbf{Evaluator expertise profiling.} While evaluators are high-performing freelancers, we did not collect fine-grained skill matrices or study how evaluator background moderates judgments or feedback utility.

\textbf{Strict completion definition.} A job is ``completed'' only if \emph{all critical and important} criteria pass---a deliberately strict standard that may under-recognize partially usable outputs and process quality (e.g., clarifying questions).

\textbf{Limited exploration of retry depth.} Most agents were evaluated only up to $k{=}2$ (with $k{=}3$ only sparsely), limiting learning-curve analyses.

\textbf{Economic modeling is illustrative.} The expected-value analysis uses observed costs and simplifying assumptions; it omits several operational frictions and risk adjustments.

\textbf{Clustering methodology.} The LLM-aided clustering pipeline (single embedding model, HAC, LLM-generated labels) may be sensitive to embedding choice, thresholds, and labeler bias.

Together, these limitations suggest the reported results are conservative estimates of achievable performance. Future work will address these gaps by adding AI-only controls, measuring evaluator reliability, and expanding the dataset for greater robustness and replicability.
\section{Conclusion}

This paper introduces UpBench, a novel benchmark that grounds agentic AI evaluation in real-world, economically verified tasks. By leveraging authentic labor-market data from the Upwork platform, UpBench captures the complexity, variability, and professional standards inherent in digital work. Unlike static or synthetic datasets, it reflects the true structure of human demand and provides a rigorous, economically meaningful foundation for measuring AI performance across diverse domains.

Beyond serving as a benchmark, UpBench establishes a continually evolving data-collection framework that integrates expert human oversight throughout every stage—from job selection and rubric construction to feedback generation and evaluation. This dynamic process ensures that the benchmark remains current with the changing nature of work and supports the study of adaptive, feedback-driven AI systems. Ultimately, UpBench represents a commitment to human–AI collaboration, offering not only a tool for measuring performance but also a testbed for understanding how human expertise and agentic intelligence can co-evolve. By embedding human judgment, feedback, and context into the benchmarking process itself, UpBench advances the goal of building AI systems that augment rather than replace human capability in the digital economy.
\section{Acknowledgements}

We gratefully acknowledge the dedicated expert freelancers -- Josue Gutierrez, Jinnah Dorothy B. Jimenez, Moshin Kamal, Inah Nicado, Charles Chukwudozie Orakwue, R. Quinones, Dr. Muhammad Sharif, and Rita Zamora --whose work made this project possible.  We would like to specifically thank Celeste Fejzo for her leadership and project management in coordinating all of our expert talent in this project.
%\intput{sections/sample}

\clearpage
% --------------------------------------------------
% Bibliography
% --------------------------------------------------
%\bibliographystyle{jmlr2e}  % or plainnat (pick one)
\bibliography{references}

@inproceedings{terzolo2025upbench,
  title     = {Towards Real-World Evaluation of Agentic Work in Freelance Marketplaces},
  author    = {Mattie Terzolo and Teng Liu and Darvin Yi and Karthik Gomadam and Lance Hasson and Ayan Sinha and Pablo Mendes and Andrew Rabinovich},
  booktitle = {Proceedings of the 39th Conference on Neural Information Processing Systems (NeurIPS 2025)},
  year      = {2025},
  organization = {Upwork Inc.}
}

@article{chen2021evaluating,
  title   = {Evaluating Large Language Models Trained on Code},
  author  = {Chen, Mark and Tworek, Jerry and Jun, Heewoo and Yuan, Qiming and et al.},
  journal = {arXiv preprint arXiv:2107.03374},
  year    = {2021},
  url     = {https://arxiv.org/abs/2107.03374}
}

@inproceedings{honeycutt2020soliciting,
  title={Soliciting human-in-the-loop user feedback for interactive machine learning reduces user trust and impressions of model accuracy},
  author={Honeycutt, Donald and Nourani, Mahsan and Ragan, Eric},
  booktitle={Proceedings of the AAAI Conference on Human Computation and Crowdsourcing},
  volume={8},
  pages={63--72},
  year={2020}
}

@article{mazeika2025remote,
  title={Remote Labor Index: Measuring AI Automation of Remote Work},
  author={Mazeika, Mantas and Gatti, Alice and Menghini, Cristina and Sehwag, Udari Madhushani and Singhal, Shivam and Orlovskiy, Yury and Basart, Steven and Sharma, Manasi and Peskoff, Denis and Lau, Elaine and others},
  journal={arXiv preprint arXiv:2510.26787},
  year={2025}
}

@inproceedings{hauptposition,
  title={Position: AI Should Not Be An Imitation Game: Centaur Evaluations},
  author={Haupt, Andreas and Brynjolfsson, Erik},
  booktitle={Forty-second International Conference on Machine Learning Position Paper Track},
  year={2025}
}

@article{patwardhan2025gdpval,
  title={GDPval: Evaluating AI Model Performance on Real-World Economically Valuable Tasks},
  author={Patwardhan, Tejal and Dias, Rachel and Proehl, Elizabeth and Kim, Grace and Wang, Michele and Watkins, Olivia and Fishman, Sim{\'o}n Posada and Aljubeh, Marwan and Thacker, Phoebe and Fauconnet, Laurance and others},
  journal={arXiv preprint arXiv:2510.04374},
  year={2025}
}

@article{vidgen2025ai,
  title={The AI Productivity Index (APEX)},
  author={Vidgen, Bertie and Fennelly, Abby and Pinnix, Evan and Mahapatra, Chirag and Richards, Zach and Bridges, Austin and Huang, Calix and Hunsberger, Ben and Zafar, Fez and Foody, Brendan and others},
  journal={arXiv preprint arXiv:2509.25721},
  year={2025}
}

@article{zhu2025establishing,
  title={Establishing Best Practices for Building Rigorous Agentic Benchmarks},
  author={Zhu, Yuxuan and Jin, Tengjun and Pruksachatkun, Yada and Zhang, Andy and Liu, Shu and Cui, Sasha and Kapoor, Sayash and Longpre, Shayne and Meng, Kevin and Weiss, Rebecca and others},
  journal={arXiv preprint arXiv:2507.02825},
  year={2025}
}

@article{liu2023agentbench,
  title={Agentbench: Evaluating llms as agents},
  author={Liu, Xiao and Yu, Hao and Zhang, Hanchen and Xu, Yifan and Lei, Xuanyu and Lai, Hanyu and Gu, Yu and Ding, Hangliang and Men, Kaiwen and Yang, Kejuan and others},
  journal={arXiv preprint arXiv:2308.03688},
  year={2023}
}

@article{kapoor2025holistic,
  title={Holistic Agent Leaderboard: The Missing Infrastructure for AI Agent Evaluation},
  author={Kapoor, Sayash and Stroebl, Benedikt and Kirgis, Peter and Nadgir, Nitya and Siegel, Zachary S and Wei, Boyi and Xue, Tianci and Chen, Ziru and Chen, Felix and Utpala, Saiteja and others},
  journal={arXiv preprint arXiv:2510.11977},
  year={2025}
}

@misc{miserendino2025swelancerfrontierllmsearn,
	title={SWE-Lancer: Can Frontier LLMs Earn \$1 Million from Real-World Freelance Software Engineering?}, 
	author={Samuel Miserendino and Michele Wang and Tejal Patwardhan and Johannes Heidecke},
	year={2025},
	eprint={2502.12115},
	archivePrefix={arXiv},
	primaryClass={cs.LG},
	url={https://arxiv.org/abs/2502.12115}, 
}

@misc{yao2023reactsynergizingreasoningacting,
	title={ReAct: Synergizing Reasoning and Acting in Language Models}, 
	author={Shunyu Yao and Jeffrey Zhao and Dian Yu and Nan Du and Izhak Shafran and Karthik Narasimhan and Yuan Cao},
	year={2023},
	eprint={2210.03629},
	archivePrefix={arXiv},
	primaryClass={cs.CL},
	url={https://arxiv.org/abs/2210.03629}, 
}

\clearpage
% --------------------------------------------------
% Appendix
% --------------------------------------------------
\appendix
\section{Instructions Given to Freelancers}\label{app:instructions}

\subsection{Instruction for Context Complete Labeling}

\begin{framed}
**Definition:** A project meets this criterion if all necessary information to complete the work is clearly provided and accessible within the job description and attachments, with no missing details or inaccessible resources.

**Requirements:**
- All necessary information is contained within attachments
- Task does not require physical presence or location-based activities  
- NO substantial follow-up questions needed to complete the work
- **Important:** Only consider job description and attachments - DO NOT evaluate deliverables for this criterion

**Common Problems to Check:**
- **Document Count:** If specific numbers mentioned, verify exact count provided (not just examples)
- **External Access:** Work requires unavailable external accounts/systems (Shopify admin, proprietary databases)
- **Missing Details:** Generic tasks lack specific company/industry/scenario details
- **Broken Links:** Referenced links or resources are inaccessible
- **Templates vs. Materials:** Attachments are examples to create, not materials to work with
- **Essential Information:** Missing product names, titles, subjects, or other completion details
\end{framed}

\subsection{Instruction for PII Labeling}

\begin{framed}
**Definition:** A project meets this criterion if all personal, company, and client identifiers are not present in the attachments or job title/description

**Requirements:**
- No personal identifying information (PII) present
- Company/organization identity is not discernible
- Job poster identity is protected
- Any remaining identifiers are appropriately anonymized
- **Important:** Only consider job description and attachments - DO NOT evaluate deliverables for this criterion
\end{framed}

\subsection{Instruction for Rubric Creation}

\begin{framed}
**Purpose:** After qualifying a project, develop 5-20 objective acceptance criteria that a client could use to evaluate whether a freelancer's work submission meets the contract requirements.

** Requirements**
Your criteria must be:
- **Objective \& Measurable:** Based on verifiable facts, not subjective opinions
- **Specific \& Concrete:** References exact deliverable requirements from project description
- **Clear \& Unambiguous:** Anyone should be able to check compliance easily
- **Quick to Verify:** Each criterion should be able to be verified in less than 1-2 minutes
- **Essential Elements:** Covers key requirements that define successful completion
- **Concise:** Should read like bullet points and be brief
**Important:** Do not include anything about requested timelines/dates or any attributes about the worker being requested (Ie ignore things about expertise in certain domain or geographic location)

*** Format Guidelines***
The acceptance criteria cell in the worksheet should contain each acceptance criteria as a line with the following structure:
```
label - criteria text
label - criteria text
...
```

** Possible values for `label`**
- `critical` - The criterion is critical to the project and must be met for the project to be accepted
- `important` - The criterion is important to the project and should be met for the project to be accepted
- `optional` - The criterion is optional and can be met for the project to be accepted
- `pitfall` - The criterion is a pitfall that must be avoided for the project to be accepted. This is different from the other types in that it is a negative attribute we want to ensure projects do not have.

** Guidelines for `criteria text`**
- Use active, definitive language ("Contains...", "Includes...", "Demonstrates...")
- Avoid subjective terms ("high-quality", "professional", "adequate")
- Leave empty if no discernible acceptance criteria exist

**Examples of Good Acceptance Criteria**
- "Deliverable contains analysis of all 12 months of data as specified in the dataset"
- "Final report includes 5 strategic recommendations with supporting rationale"
- "All visualizations are properly labeled with titles, axis labels, and data source citations"
- "Must contain between 90-120 words"

** Examples of Poor Acceptance Criteria**
- "Work is of professional quality" *(subjective)*
- "Analysis is thorough and insightful" *(unmeasurable)*
- "Analysis must be completed by end of the week" *(timeline-related)*
- "Must be completed by U.S based worker" *(worker-related)*
\end{framed}

\subsection{Instruction for Deliverable Evaluation}

\begin{framed}
** Instructions for the Evaluation Task**

You are evaluating submitted work deliverables against objective acceptance criteria to determine whether they meet the success criteria established for each project.

** Main Task: Submission Evaluation**

Evaluate submitted work deliverables against the acceptance criteria established during project qualification to determine if submissions meet contractual requirements.

** Evaluation Criteria Application**

For each acceptance criterion provided, you must evaluate whether the submission meets that specific requirement:

**pass**: The submission clearly and demonstrably meets the criterion

**skip**: The criterion cannot be properly evaluated or is not applicable
Common reasons to skip: 
- Criterion not relevant to what was requested in the job post
- Do not have the tools or skills to evaluate the criterion

**fail**: The submission does not meet the criterion

**Important Guidelines**:
- Base judgments solely on objective, verifiable facts
- Reference specific deliverable content when making assessments
- Use the exact acceptance criteria provided - do not modify or interpret them
- Each criterion should be evaluated independently
- Focus on contract compliance, not work quality beyond the stated criteria

\end{framed}
\section{Data Variable Details}

\subsection{Database Schema Overview}
The database consists of six main tables: \texttt{projects}, \texttt{qualifications}, \texttt{submissions}, \texttt{evaluations}, \texttt{evaluators}, and \texttt{agents}. Each table records a different part of the benchmarking workflow.

\begin{table}[h]
\centering
\caption{Project Metadata (\texttt{projects} table)}
\begin{tabular}{p{3cm} p{9cm}}
\hline
\textbf{Variable} & \textbf{Description} \\
\hline
\texttt{contract\_id} & Unique identifier for each job or contract. \\
\texttt{start\_ts}, \texttt{end\_ts} & Start and end timestamps for the contract. \\
\texttt{job\_title}, \texttt{job\_description} & Title and textual description of the job. \\
\texttt{job\_amount}, \texttt{contract\_amount}, \texttt{milestone\_amount} & Alternative payment amount fields. \\
\texttt{expertise\_tier} & Job difficulty or expertise level. \\
\texttt{category}, \texttt{subcategory}, \texttt{subsubcategory} & Hierarchical classification of job sectors. \\
\texttt{num\_openings}, \texttt{num\_milestones}, \texttt{automation\_desire\_rating} & Numeric meta fields describing milestones and automation potential. \\
\texttt{onet\_tasks} & O*Net task references for the job. \\
\texttt{attachments} & JSON list of attachments linked to the project. \\
\hline
\end{tabular}
\end{table}

\begin{table}[h!]
\centering
\caption{Qualification Data (\texttt{qualifications} table)}
\begin{tabularx}{\textwidth}{p{4cm} X}
\hline
\textbf{Variable} & \textbf{Description} \\
\hline
\texttt{\detokenize{qualification_id}} & Unique identifier for each qualification event. \\
\texttt{\detokenize{contract_id, evaluator_id}} & Job and evaluator references. \\
\texttt{\detokenize{qualification_ts}} & Timestamp of the qualification event. \\
\texttt{\detokenize{criterion_1--3_}}\\
\texttt{\detokenize{judgment/reasoning}} & Ratings for context completeness, deliverable quality, and PII absence. \\
\texttt{\detokenize{criterion_4, criterion_5}} & Additional or experimental criteria. \\
\texttt{\detokenize{acceptance_criteria}} & JSON rubric defining evaluation standards. \\
\texttt{\detokenize{is_successful_run}} & Indicates whether the qualification process completed successfully. \\
\hline
\end{tabularx}
\end{table}

\begin{table}
\centering
\caption{Submission Data (\texttt{submissions} table)}
\begin{tabular}{p{3cm} p{9cm}}
\hline
\textbf{Variable} & \textbf{Description} \\
\hline
\texttt{submission\_id} & Unique identifier for each submission. \\
\texttt{contract\_id}, \texttt{agent\_id} & References to the job and submitting agent. \\
\texttt{submission\_ts} & Timestamp of the submission. \\
\texttt{deliverables}, \texttt{notes} & Submitted content and any accompanying notes. \\
\texttt{k} & Attempt number (e.g., first, second, etc.). \\
\texttt{duration} & Duration of the attempt in seconds. \\
\texttt{is\_successful\_run} & Indicates if the submission was successfully processed. \\
\hline
\end{tabular}
\end{table}

\begin{table}
\centering
\caption{Evaluation Data (\texttt{evaluations} table)}
\begin{tabular}{p{3.5cm} p{10cm}}
\hline
\textbf{Variable} & \textbf{Description} \\
\hline
\texttt{evaluation\_id} & Unique identifier for each evaluation event. \\
\texttt{contract\_id}, \texttt{submission\_id}, \texttt{evaluator\_id} & Links to job, submission, and evaluator records. \\
\texttt{submission\_feedback} & JSON rubric-based feedback for the submission. \\
\texttt{judgement} & Overall evaluation outcome (pass/fail). \\
\texttt{is\_successful\_run} & Indicates if the evaluation process completed successfully. \\
\hline
\end{tabular}
\end{table}

\begin{table}
\centering
\caption{Evaluator and Agent Metadata (\texttt{evaluators} and \texttt{agents} tables)}
\begin{tabular}{p{3cm} p{9cm}}
\hline
\textbf{Variable} & \textbf{Description} \\
\hline
\texttt{evaluator\_id}, \texttt{agent\_id} & Unique identifiers for evaluator and worker agents. \\
\texttt{registration\_ts} & Timestamp of registration. \\
\texttt{is\_human} & Boolean flag for human or AI status. \\
\texttt{name} & Name of the evaluator or agent (e.g., \texttt{Rita}, \texttt{darvin-ms-eval-gpt-5}). \\
\hline
\end{tabular}
\end{table}

\section{Metrics}

\paragraph{Lift Calculations for Success Score}
\subsection*{1. Absolute Lift (percentage points)}
\[
\Delta p = p_{\text{overall}} - p_1 = \frac{A + B}{N} - \frac{A}{N} = \frac{B}{N}
\]
\textbf{Interpretation:} Increase in overall pass rate (in percentage points) due to adding HITL.

\subsection*{2. Relative Lift (percent improvement)}
\[
\text{Relative Lift} = \frac{p_{\text{overall}}}{p_1} - 1 = \frac{B}{A}
\]
\textbf{Interpretation:} Percent improvement in success rate compared to AI-only baseline.

\subsection*{3. Rescue Rate (conditional on failures)}
\[
\text{Rescue Rate} = p_{\text{H|F}} = \frac{B}{M}
\]
\textbf{Interpretation:} Among jobs that failed initially \textbf{and} received a HITL re-attempt, 
the fraction successfully ``rescued'' by HITL.\\

% \subsection*{Notes}
% \begin{itemize}
%     \item The denominator $N$ stays fixed across all comparisons—it represents the same set of original jobs.
%     \item Use the \textbf{absolute lift} to communicate system-level improvement, 
%     and the \textbf{rescue rate} to show the effectiveness of HITL interventions.
% \end{itemize}

\paragraph{Lift Calculations for Rubric Score}

% \subsection*{1. Absolute Lift (percentage points)}
% \[
% \Delta \bar{R} \;=\; \bar{R}_{\text{overall}} - \bar{R}_1,
% \]

% \subsection*{2. Relative Lift (percent improvement)}
% Then the HITL-conditional lifts are
% $$
% \text{RelLift}_{H \mid F} \;=\; \frac{\bar{R}_{2 \mid M}}{\bar{R}_{1 \mid M}} - 1
% $$
% These are the graded analogues of the rescue-rate perspective: they quantify how much quality improves among the jobs that actually received HITL.

% \paragraph{Rubric lift metrics (job-weighted).}
% We report two lifts that are cohort-consistent with the pass-rate framework.

\paragraph{1. Absolute overall lift (rubric points)}
\[
\Delta \bar{R} \;=\; \bar{R}_{\text{overall}} - \bar{R}_1,
\]
where
\(
\bar{R}_1 = \tfrac{1}{|N|}\sum_{j\in N} R_{j,1}
\)
and
\(
\bar{R}_{\text{overall}} = \tfrac{1}{|N|}\sum_{j\in N} R_{j,\text{final}}
\)
with
\(
R_{j,\text{final}}=R_{j,1}\ \text{if } j\in A,\ 
R_{j,2}\ \text{if } j\in M,\ 
R_{j,1}\ \text{if } j\in F\setminus M.
\)
(Units: rubric points on $[0,1]$. If reported in percentage points, multiply by $100$.)

\paragraph{2. Relative lift among HITL jobs (percent improvement)}
Define the baseline and post-HITL means on the same failed-and-retried set $M$:
\[
\bar{R}_{1\mid M} = \frac{1}{|M|}\sum_{j\in M} R_{j,1},
\qquad
\bar{R}_{2\mid M} = \frac{1}{|M|}\sum_{j\in M} R_{j,2}.
\]
Then the HITL-conditional relative lift is
\[
\text{RelLift}_{H \mid F} \;=\; \frac{\bar{R}_{2 \mid M}}{\bar{R}_{1 \mid M}} - 1.
\]
This is the graded analogue of the rescue-rate view: it measures the percent improvement in rubric quality among the jobs that actually received HITL. (If $|M|=0$ or $\bar{R}_{1\mid M}=0$, this quantity is undefined.)
\section{Modeling}\label{app:modelling}

\subsection{Modified \texorpdfstring{\texttt{pass@k}}{pass@k} for Real-World Agent Jobs}
\label{app:passk}

We evaluate an AI Agent's ability to complete real jobs using a modification of the standard \texttt{pass@k} metric for functional correctness of code generation \citep{chen2021evaluating}. For each task \(i\) in an evaluation suite \(\mathcal{P}\), we generate \(n_i \ge k\) independent attempts and observe \(c_i \le n_i\) successful attempts (i.e., attempts that pass validation/tests). The unbiased estimator of \texttt{pass@k} for task \(i\) is
\begin{equation}
\widehat{\mathrm{pass}}_i(k)
\;=\;
1 \;-\; \frac{\binom{n_i - c_i}{k}}{\binom{n_i}{k}},
\qquad\text{with the convention that } \widehat{\mathrm{pass}}_i(k)=1 \text{ if } n_i - c_i < k,
\label{eq:passk_unbiased}
\end{equation}
and the aggregate score averages this quantity over tasks:
\begin{equation}
\widehat{\mathrm{pass}}(k)
\;=\;
\frac{1}{|\mathcal{P}|}
\sum_{i \in \mathcal{P}} \widehat{\mathrm{pass}}_i(k).
\label{eq:passk_dataset}
\end{equation}
Equation~\eqref{eq:passk_unbiased} is preferred over the naive estimator \(1-(1-\hat{p})^k\) (with \(\hat{p}=c_i/n_i\)), which is biased \citep[App.~A]{chen2021evaluating}. In practice, to avoid numerical instability when evaluating ratios of binomial coefficients, we compute \eqref{eq:passk_unbiased} using the product identity
\begin{equation}
\widehat{\mathrm{pass}}_i(k)
\;=\;
1 \;-\; \prod_{j=n_i - c_i + 1}^{n_i}\!\left(1 - \frac{k}{j}\right),
\quad\text{and if } n_i - c_i < k \text{ then } \widehat{\mathrm{pass}}_i(k)=1,
\label{eq:passk_product}
\end{equation}
which matches the stable implementation recommended in \citet{chen2021evaluating}.

\paragraph{Motivation for a ceiling parameter.}
In the original formulation, \(\widehat{\mathrm{pass}}(k)\) can approach \(1\) as the number of allowed attempts \(k\) grows, provided at least one correct attempt exists among the samples for each task. In real job settings, however, exogenous constraints can preclude perfect success even with many retries. Examples include:
\begin{itemize}
  \item limited or missing access to required external tools or data sources;
  \item runtime, budget, or rate-limit constraints on the agent;
  \item safety, policy, or environment restrictions that prevent specific actions;
  \item agent-specific limitations (e.g., weak tool-use skills) that systematically reduce feasibility.
\end{itemize}

\paragraph{Our modification.}
To reflect these constraints, we introduce an agent- and suite-specific ceiling parameter \(p_{\max}\in[0,1]\) that caps the attainable pass rate. Our modified metric is
\begin{equation}
\widehat{\mathrm{pass}}^{(\mathrm{agent})}(k)
\;=\;
p_{\max}\;\widehat{\mathrm{pass}}(k)
\;=\;
\frac{p_{\max}}{|\mathcal{P}|}\sum_{i \in \mathcal{P}}
\left(
1 - \frac{\binom{n_i - c_i}{k}}{\binom{n_i}{k}}
\right).
\label{eq:passk_capped}
\end{equation}
This multiplicative ceiling makes the asymptotic behavior explicit:
\(\lim_{k\to\infty}\widehat{\mathrm{pass}}^{(\mathrm{agent})}(k)=p_{\max}\),
rather than \(1\). Intuitively, \(p_{\max}\) summarizes the fraction of tasks in the suite that are \emph{operationally feasible} for the agent under its tool set, policies, and resource limits. We treat \(p_{\max}\) as agent-specific, with the expectation that stronger agents (e.g., better models and/or more capable tool stacks) achieve larger \(p_{\max}\).

\paragraph{Modeling choices.}
In this work we adopt the following choices:
\begin{enumerate}
  \item Use the unbiased estimator in \eqref{eq:passk_unbiased} with the stable computation \eqref{eq:passk_product} \citep{chen2021evaluating}.
  \item Interpret \(k\) as the per-task retry budget consistent with real deployment constraints.
  \item Average over tasks as in \eqref{eq:passk_dataset}.
  \item Apply a single \(p_{\max}\) per agent and evaluation suite as in \eqref{eq:passk_capped} to capture suite-level feasibility constraints; we expect more capable agents to have higher \(p_{\max}\).
\end{enumerate}
An alternative (not used here) would be to allow task-dependent ceilings \(p_{\max,i}\) to encode known feasibility attributes for specific tasks; in that case \(p_{\max}\) in \eqref{eq:passk_capped} is replaced by \(p_{\max,i}\) inside the sum.

\subsection{Model Fitting}

With $n$, $c$, and $p_{\text{max}}$, equation \ref{eq:passk_capped} has 3 degrees of freedom.  In our collected data, we have calculated pass rates at k=1 and k=2 for 3 of our agents and a k=3 (2nd HITL re-attempt) for one of our agents.  Just to get rough modeling numbers, we fit equation \ref{eq:passk_capped} on our 3 data point agent to find an estimate for $p_{\text{max}}$.  We share this value for the other two agents' model but fit $n$ and $c$ separately.  We use this to generate a rough prediction of the other two agents' performance at k=3.

Through this process, we find a fit value of $p_{\text{max}} = 0.57$.

\subsection{Limitations}\label{app:passk_limits}

While the capped \texttt{pass@k} in Eq.~\eqref{eq:passk_capped} is a useful abstraction for real‑world agent performance, our current fit has several important limitations.

\paragraph{(1) Data scarcity and identifiability.}
We fit three free parameters \(\,(n, c, p_{\max})\,\) with effectively three observed points (aggregate \(\mathrm{pass}@k\) at \(k\!\in\!\{1,2\}\) for all agents and \(k\!=\!3\) for one agent). This saturated setting is highly ill‑conditioned: small measurement noise can lead to large shifts in the fitted parameters, and multiple \((n,c,p_{\max})\) triplets can produce near‑indistinguishable curves over \(k\in\{1,2,3\}\). As a result, the \(k\!=\!3\) predictions for the two agents without \(k\!=\!3\) observations should be treated as high‑variance extrapolations rather than precise forecasts. Moreover, because we fit on aggregated pass rates, \(n\) and \(c\) should be interpreted as \emph{effective} parameters rather than literal per‑task sample counts, and heterogeneity across tasks can further blur their meaning. A larger dataset—with more tasks, more \(k\) levels (e.g., \(k\geq 3\) for all agents), and repeated measurements—would materially reduce variance and improve parameter identifiability.

\paragraph{(2) Shared ceiling parameter \(p_{\max}\).}
We transfer a single fitted ceiling (\(p_{\max}=0.57\)) from the agent with \(k\!=\!3\) data to the other agents. In practice, \(p_{\max}\) is induced by constraints that are \emph{agent–suite–environment} specific (tool availability, rate limits, budgets, safety policies, orchestration quality). Assuming a common ceiling across agents is unlikely to hold and can bias relative comparisons: stronger agents (or richer tool stacks) may legitimately attain higher ceilings, while agents with tighter constraints may have lower ceilings. Future analyses should allow \(p_{\max}\) to vary at least by agent (and potentially by evaluation suite), with partial pooling or hierarchical priors to share signal without forcing equality.

\paragraph{(3) Non‑IID re‑attempts and feedback dependence.}
The standard \texttt{pass@k} estimator presumes attempts are i.i.d.\ samples from the same proposal distribution. Our setting explicitly violates this assumption: later attempts are constructed using submission feedback and (for \(k\!=\!3\)) a human‑in‑the‑loop re‑attempt. These dependencies typically \emph{increase} success probabilities relative to i.i.d.\ sampling and break the unbiasedness guarantees of the classical estimator. Consequently, our reported curve is best interpreted as a \emph{policy‑dependent} “pass‑within‑\(k\)” under our specific feedback protocol, not as the canonical i.i.d.\ \texttt{pass@k}. Direct comparison to literature values computed under i.i.d.\ sampling may therefore be misleading.

\paragraph{Implications and recommended mitigations.}
To address the above, we plan to (i) expand data collection to include more \(k\) levels per agent and more tasks, (ii) report uncertainty (e.g., bootstrap across tasks or Bayesian posteriors) for \(p_{\max}\) and the predicted \(\mathrm{pass}@k\), (iii) estimate \(p_{\max}\) per agent (and/or per suite) with partial pooling rather than sharing a single value, and (iv) complement \texttt{pass@k} with a policy‑aware sequential metric (e.g., empirical “pass within \(k\)” under our workflow, or a simple hazard/learning‑curve model) to faithfully capture feedback‑driven retries.

\end{document}